\documentclass[journal]{IEEEtran}
\usepackage{amsmath,graphicx}
\usepackage{comment}
\usepackage{float}
\usepackage{algorithm}
\usepackage{algorithmicx}
\usepackage{algpseudocode}
\usepackage{caption}
\usepackage{subcaption}
\usepackage{comment}
\usepackage{adjustbox}
\usepackage{amsfonts} 
\usepackage{multirow}
\usepackage{nicematrix}
\usepackage{booktabs}
\usepackage{lscape}
\usepackage{amsthm}
\newtheorem{proposition}{Proposition}
\DeclareMathOperator*{\minimize}{minimize}
\DeclareMathOperator*{\maximize}{maximize}
\DeclareMathOperator*{\argmin}{argmin}

\title{Learning Optimal Graph Filters for Clustering of Attributed Graphs}
\author{Meiby~Ortiz-Bouza,~\IEEEmembership{}
        Selin~Aviyente~\IEEEmembership{}
\thanks{ M. Ortiz-Bouza and S. Aviyente were with the Department
of Electrical and Computer Engineering, Michigan State University, East Lansing,
MI, 48824.\protect\\
E-mail: ortizbou@msu.edu, aviyente@egr.msu.edu
}
\thanks{This work was supported in part by the National Science Foundation under  CCF-2211645.}}

\newcommand{\matr}[1]{\mathbf{#1}}
\begin{document}
%
\maketitle
%
\begin{abstract}
        Many real-world systems can be represented as graphs where the different entities in the system are presented by nodes and their interactions by edges. An important task in studying large datasets with graphical structure is graph clustering. While there has been a lot of work on graph clustering using the connectivity between the nodes, many real-world networks also have node attributes. Clustering attributed graphs requires joint modeling of graph structure and node attributes. Recent work has focused on combining these two complementary sources of information through graph convolutional networks and graph filtering. However, these methods are mostly limited to lowpass filtering and do not explicitly learn the filter parameters for the clustering task.  In this paper, we introduce a graph signal processing based approach, where we learn the parameters of Finite Impulse Response (FIR) and Autoregressive Moving Average (ARMA)  graph filters optimized for clustering.         The proposed approach is formulated as a two-step iterative optimization problem, focusing on learning interpretable graph filters that are optimal for the given data and that maximize the separation between different clusters. The proposed approach is evaluated on attributed networks and compared to the state-of-the-art methods.
\end{abstract}

\section{Introduction}
\label{sec:intro}
Many real-world systems with relational data such as social interactions, citation and co-author relationships, and biological systems are represented as networks \cite{barabasi2013network}. An important aspect of analyzing networks is the discovery of communities \cite{girvan2002community} which allow us to identify groups of functionally related objects and the interactions between them \cite{leskovec2012learning,ahn2010link}.

In most real-world networks, both the node attributes and graph connectivity are available.  
For example, known properties of proteins,
users’ social network profiles, or authors’ publication histories
may tell us which objects are similar, and to which communities  they may belong. Similarly, 
 the set of edges between the
nodes, such as the friendship relationships between users, interactions between proteins, and the collaboration between authors, can help identify groups based on connectivity. Classical data clustering methods such as $k$-means assign class labels based on only attribute similarity \cite{jain2010data}. On
the other hand, community detection algorithms find
groups of nodes that are densely connected \cite{fortunato2010community,tremblay2014graph,mucha_community_2010},  ignoring node attributes. Employing just one of these two sources of information can result in the algorithm failing to account for important structures in the data. For instance, while it would be hard to determine the community membership of a sparsely connected node by solely relying on the network's connectivity, attributes may help reveal the community affiliation. Conversely, the network may suggest that two nodes belong to the same community, even if one lacks attribute information. Hence, it is crucial to take both information sources into account and view network communities as clusters of closely linked nodes that also share common attributes.
\vspace{-0.1in}
\subsection{Related Work}
In recent years, several methods have been proposed for attributed graph clustering by combining the node attributes and link information \cite{leskovec2012learning,jia2017node,zhang2019attributed,alinezhad2020community,yang2013community}. The first class of methods for attributed graph clustering focuses on combining link and node information by formulating an objective function that integrates the two types of similarity: adjacency matrix that captures link information and the similarity matrix that quantifies the affinity between the attributes  \cite{lu2022community,alinezhad2020community,malhotra2021modified}. These methods are indirect in the sense that they rely on the construction of an arbitrary similarity matrix from the node attributes.
The second class of methods incorporates the graph structure and node attributes simultaneously into the community detection framework benefiting from the representation capability of graph neural networks (GNNs) \cite{yue2022survey}. The goal of these methods is to encode nodes in the graph with the neural networks and assign node labels.   For example, 
methods such as graph autoencoders (GAE) \cite{kipf2016semi}, variational GAE (VGAE) \cite{kipf2016variational}, adversarially regularized graph autoencoder (ARGA), adversarially regularized variational graph autoencoder (ARVGA) \cite{pan2018adversarially} and marginalized graph autoencoder for graph clustering (MGAE) \cite{wang2017mgae} have demonstrated state-of-the-art performance on attributed graph clustering. Although these methods achieve promising performance, they are not designed for the specific clustering performance, i.e., the network parameters are optimized to minimize the reconstruction error rather than maximizing the separation between different clusters.  Moreover, in these methods, each convolutional layer is coupled with
a projection layer, making it difficult to stack many layers
and train a deep model. Thus, they only take into account neighbors of each node in two or three hops away and hence may be inadequate to capture global cluster structure of large graphs. Finally, these methods lack interpretability as they do not explicitly show the relationship between the learned models and the structure of the graph. 

In order to take advantage of graph convolutional features while addressing the shortcomings of GNN-based methods, the last class of methods rely on graph signal processing (GSP) and in particular spectral graph filtering
\cite{zhang2019attributed,li2019label,xie2022graphhop,kang2022fine,wu2022beyond}. Over the past decade graph filters \cite{isufi2022graph} have played a key role in different signal processing and machine learning tasks such as graph signal
denoising \cite{deutsch2018robust,onuki2016graph}, smoothing \cite{zhang2008graph}, classification \cite{sandryhaila2013classification,chen2014semi}, sampling
\cite{anis2016efficient}, recovery \cite{chen2015signal}, and graph clustering \cite{tremblay2016compressive}. Different types of graph filter structures, including FIR graph filters \cite{segarra2017optimal,shuman2018distributed}, ARMA filters \cite{isufi2016autoregressive}, 
graph filter banks \cite{narang2012perfect}, and graph wavelets \cite{hammond2011wavelets,narang2013compact} have been considered.  For the task of graph node classification, two different classes of approaches have been developed. The first class of methods is semi-supervised learning on graphs, and graph filters are
used to weigh and propagate the label information of multi-hop neighbors to the unknown nodes. This problem is formulated as an optimization problem, where the filter parameters are estimated to minimize the error between the estimated labels and the true labels on the labeled nodes with regularization on the filter parameters or the output, e.g., smooth label variation \cite{chen2014semi,fan2022graph,berberidis2018adaptive}. These works only consider the node label and graph connectivity information but do not necessarily address the issue of unsupervised clustering of attributed graphs. In the realm of unsupervised learning, GSP techniques have been used to address clustering and community mining problems. In \cite{tremblay2014graph}, spectral graph wavelets are utilized to develop a fast, multiscale community mining protocol. In \cite{tremblay2016compressive}, graph-spectral
filtering of random graph signals is used to construct feature vectors for each vertex so that the distances between vertices based on these feature vectors resemble those based on standard spectral clustering feature vectors. More recently, \cite{donnat2018learning} uses spectral graph wavelets to learn structural embeddings that help identify vertices that have similar structural roles in the network. In all of these cases, the problem of community detection is only addressed for either graphs without attributes or regular data clustering without graph structure. Moreover, the graph filter parameters are fixed. More recently, adaptive graph convolution (AGC) \cite{zhang2019attributed} was proposed for attributed graph clustering. Instead of stacking layers as in GCN, 
a $k$-order graph convolution that acts as a low-pass graph filter on node features is proposed to obtain smooth feature representations followed by spectral clustering on the learned features. In \cite{kang2022fine}, this approach is further refined by learning the best similarity graph from the filtered features rather than constructing a graph using pre-determined similarity metrics.  While these methods are intuitive and provide some interpretability to the node features, the filters are always low-pass, ignoring the useful information in higher frequency bands \cite{wu2022beyond},  and the filter shape is not optimized for the particular data. Moreover, the two steps of the algorithm, i.e., filtering and clustering, are completely decoupled from each other. Thus, there is no guarantee that the extracted features are optimal for clustering. 
\vspace{-0.1in} 
\subsection{Contributions}
In this paper, we address the shortcomings of the existing methods by proposing a graph filtering based method, GraFiCA,  for community detection in attributed networks. A cost function quantifying the separability of the filtered attributes is proposed and a general framework for learning the parameters of both FIR and ARMA filters is introduced.  The proposed approach is formulated as a two-step alternating minimization problem, where the first step learns the optimal graph partitioning for the given node attributes while the second step learns the optimal graph filter parameters.

The main contributions of the proposed framework are as follows: 
\begin{itemize}
\item The proposed framework is the first that addresses the problem of parametric graph filter design in the form of both FIR and IIR filters for the purpose of attributed graph clustering. While there has been prior work in graph filter design for denoising or graph signal recovery \cite{routtenberg2021non}, GraFiCA is the first unsupervised approach that learns the parameters of the filters for the purpose of clustering.
    \item  The filters learned by the proposed approach are not limited to low-pass filters as the structure of the filter is determined directly by the data. The filters take into account the useful information in middle and high frequency bands, i.e., higher-order neighborhoods,  providing interpretability to the learned filters.
    \item  GraFiCA uses a cost function that quantifies the discriminability between different classes unlike GCN-based approaches where the loss function is usually the reconstruction error. Thus, the filter coefficients are updated at each step of the algorithm to ensure that the smoothed node attributes are representative of the node assignments.
\end{itemize}

\section{Preliminaries}
\label{sec:bckgnd}
\subsection{Graphs}

Let $\mathcal{G}=(V, E, \matr{A})$ be a graph where $V$ is the node set with $|V|=N$, $E$ is the edge set and $\matr{A}\in \mathbb{R}^{N\times N}$ is the adjacency matrix. The graph Laplacian is given by $\boldsymbol{\matr{L}}=\matr{D-A}$, where $\matr{D}$ is the diagonal degree matrix defined as $D_{ii}=\sum_{j} A_{ij}$.
The normalized Laplacian matrix $\matr{L}_{n}$ is defined as $\matr{L}_{n}=\matr{D}^{-1/2}(\matr{D-A})\matr{D}^{-1/2}=\matr{I}_N-\matr{D}^{-1/2}\matr{A}\matr{D}^{-1/2}=\matr{I}_N-\matr{A}_{n}$, where $\matr{I}_N$ is the identity matrix of size $N$ and $\matr{A}_{n}$ is the normalized adjacency matrix. The spectrum of $\matr{L}_{n}$ is composed of the diagonal matrix of the eigenvalues, $\boldsymbol\Lambda = \text{diag}(\lambda_1,\ldots,\lambda_N)$ with $\lambda_1 \leq \lambda_2 \leq \ldots \leq \lambda_N$, and the eigenvector matrix  $\matr{U}=[u_1|u_2|\ldots|u_N]$ such that $\matr{L}_{n}=\matr{U}\matr{\Lambda}\matr{U}^\top$
\cite{chung1997spectral}. 

\subsection{Spectral Clustering}
\label{sec:datacluster}
Given data with $N$ samples, $(x_{1},x_{2},\ldots,x_{N})$, the intuition of clustering is to partition the data points based on their similarities. Let $s_{ij}=s(x_{i},x_{j})$ be the pairwise similarity between two data objects quantified by some similarity function which is symmetric and nonnegative. Given the similarity graph with an adjacency matrix $\matr{W}$, the problem of data clustering can be restated into graph partitioning such that the edges between different groups have a very low weight and the edges within a group have a high weight. For finding $K$ clusters, minimizing the cut of this graph consists of finding a partition $\{\mathcal{C}_{1},\mathcal{C}_{2},\ldots,\mathcal{C}_{K}\}$ which minimizes:
\begin{equation}
    \text{Cut}(\mathcal{C}_{1},\mathcal{C}_{2},\ldots,\mathcal{C}_{K})=\frac{1}{2}\sum_{k=1}^{K} \text{links}(\mathcal{C}_{k},V\setminus \mathcal{C}_k),
  \label{eq:cut}  
\end{equation}
where $\text{links}(\mathcal{C}_k,V\setminus \mathcal{C}_k)=\sum_{i\in \mathcal{C}_k,j\notin \mathcal{C}_k} W_{ij}$. In practice, minimizing the cut results in clusters that are unbalanced. In order to address this issue, two different variations of the cut definition, RatioCut (RCut) \cite{hagen1992new} and NormalizedCut (Ncut) \cite{shi2000normalized}, have been proposed: 
    \begin{equation}
    \begin{split}
\text{RCut}(\mathcal{C}_{1},\mathcal{C}_{2},\ldots,\mathcal{C}_{K})=\sum_{k=1}^{K}\frac{\text{links}(\mathcal{C}_{k},V\setminus \mathcal{C}_k)}{|\mathcal{C}_{k}|},\\
\text{NCut}(\mathcal{C}_{1},\mathcal{C}_{2},\ldots,\mathcal{C}_{K})=\sum_{k=1}^{K}\frac{\text{links}(\mathcal{C}_{k},V\setminus \mathcal{C}_k)}{\text{vol}(\mathcal{C}_{k})},
\end{split}
\label{eq:cuts}
    \end{equation}
    
\noindent where $|\mathcal{C}_{k}|$ is the number of nodes in $\mathcal{C}_{k}$ and $\text{vol}(\mathcal{C}_{k})$ is the total degree of all nodes in $\mathcal{C}_{k}$.

Similar to minimizing the cut metrics, one can determine the partition by maximizing the corresponding association metrics \cite{dhillon2004unified}. For example, the normalized association is defined as:
\begin{equation}
    \text{NAssoc}(\mathcal{C}_{1},\mathcal{C}_{2},\ldots,\mathcal{C}_{K})=\sum_{k=1}^{K}\frac{\text{links}(\mathcal{C}_{k},\mathcal{C}_{k})}{\text{vol}(\mathcal{C}_{k})}.
    \label{eq:assoc}
\end{equation}

Minimizing or maximizing these cost functions are NP hard. Spectral clustering is a way to solve relaxed versions of these problems \cite{von2007tutorial}. 
Introducing a cluster indicator vector for cluster $k$, i.e. $z_{k_i}=1$ when node $i$ belongs to cluster $k$, we can set the matrix $\matr{Z} \in \mathbb{R}^{N\times K}$ as the matrix whose columns correspond to the $K$ indicator vectors. The normalized association can then be rewritten as $\sum_{k=1}^{K}\frac{\text{links}(\mathcal{C}_k,\mathcal{C}_k)}{\text{vol}(\mathcal{C}_k)}=\sum_{k=1}^{K}\frac{z_k\matr{W}z_k^\top}{z_k\matr{D}z_k^\top}=\sum_{k=1}^{K} \tilde{z_k}\matr{W}\tilde{z_k}^\top $ with $\tilde{z_k} = z_k/(z_k^\top\matr{D}z_k)^{1/2}$. 
This can be rewritten as the following optimization problem
\begin{equation}
\underset{\matr{\bar{Z}},\matr{\bar{Z}}^\top\matr{\bar{Z}}=\matr{I}}{\maximize} \hspace{0.1in}\text{tr}(\matr{\bar{Z}}^\top \matr{D}^{-1/2}\matr{W}\matr{D}^{-1/2}\matr{\bar{Z}}),
    \label{eq:traceNassoc}
\end{equation}
\noindent where $\matr{\bar{Z}} = \matr{D}^{1/2}\matr{\tilde{Z}}$. 
This is the standard form of a trace maximization problem,
and the Rayleigh-Ritz theorem  states that the solution
is given by choosing $\matr{\bar{Z}}$ as the matrix which contains the $K$ eigenvectors corresponding to the largest eigenvalues of $\matr{D}^{-1/2}\matr{W}\matr{D}^{-1/2}$ as columns.

Similar to clustering data points, one of the most fundamental tasks in analyzing large-scale networks is community detection \cite{fortunato2010community}. Community detection is equivalent to graph partitioning and aims to uncover groups of nodes with higher connectivity amongst themselves compared to the rest of the network. 
Given a graph with adjacency matrix $\matr{A}$, which represents the graph connectivity information, one can find the communities using spectral clustering by either minimizing the NCut or maximizing the NAssoc of the graph yielding
\begin{equation}
\underset{\matr{\bar{Z}},\matr{\bar{Z}}^\top\matr{\bar{Z}}=\matr{I}}{\maximize} \hspace{0.1in}\text{tr}(\matr{\bar{Z}}^\top \matr{D}^{-1/2}\matr{A}\matr{D}^{-1/2}\matr{\bar{Z}}).
    \label{eq:traceNassocA}
\end{equation}
\vspace{-0.2in}
\subsection{Graph Filtering}
In graph signal processing, two fundamental filter types, Finite Impulse Response (FIR) and Autoregressive Moving Average (ARMA) \cite{isufi2016autoregressive} graph filters, are considered \cite{isufi2022graph}. FIR polynomial graph filter is described as the linear operator
\begin{equation*}
    \mathcal{H}(\matr{L})=\sum_{t=0}^{T-1} h_t\matr{L}^t=\matr{U}(\sum_{t=0}^{T-1} h_t\boldsymbol{\Lambda}^t)\matr{U}^\top,
\end{equation*}
\vspace{-0.1in}
\noindent where $T$ is the filter order and $h_t$'s are the coefficients. 
On the other hand, an ARMA filter is defined as
\begin{equation*}
    \mathcal{H}(\matr{L})=\matr{U}\frac{\sum_{t=0}^{T-1} a_t\boldsymbol{\Lambda}^t}{\matr{I}+\sum_{q=1}^{Q-1} b_q\boldsymbol{\Lambda}^q}\matr{U}^\top,
\end{equation*}
\noindent where $(T,Q)$ are the filter orders and $a_t$'s and $b_q$'s are the filter coefficients. Note that the FIR polynomial filter is a special case of ARMA filter for $Q=1$.

Signals defined on the nodes of an attributed graph can be represented as a matrix $\matr{F}\in \mathbb{R}^{N\times P}$, where $P$ is the number of attributes for each node. The filtered graph signal $\matr{\tilde{F}}$ is obtained as $\matr{\tilde{F}}= \mathcal{H}(\matr{L})\matr{F}=\matr{U}\mathcal{H}(\boldsymbol{\Lambda})\matr{U}^\top\matr{F}$, where $\mathcal{H}(\boldsymbol\Lambda) = \text{diag}(\mathcal{H}(\lambda_1),\ldots, \mathcal{H}(\lambda_N))$ is the frequency response of the graph filter.
\vspace{-0.15in}
\section{Graph Filtering for Clustering Attributed Graphs (G\MakeLowercase{ra}F\MakeLowercase{i}CA)}
\label{sec:propmeth}
\begin{figure*}
    \centering
    \includegraphics[width=\linewidth]{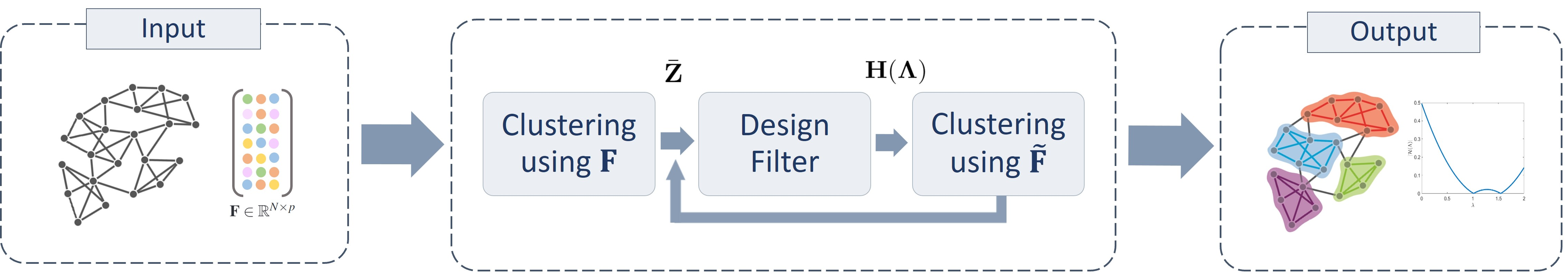}
    \caption{Framework of the proposed method}
    \label{fig:fram}
\end{figure*}

In this paper, we propose to learn the optimal graph filter such that the within-cluster association of the filtered attributes is minimized while the between-class distance of the filtered attributes is maximized. Using an alternating minimization approach, in the first step, given the node attributes, we find the best cluster assignment to minimize the clustering cost function. In the second step, the graph partition is fixed and the cost function is optimized with respect to the filter coefficients. In this manner, the resulting graph filters are optimized for the clustering task (see Fig. \ref{fig:fram} for an overview of the method). This results in an attributed graph clustering method that takes into account both the topology and the node attributes.

In this section, we will first introduce the general problem formulation and the corresponding optimization problem. We will then present solutions for the optimal filter design for two different filter types; FIR and ARMA.
\vspace{-0.15in}
\subsection{Problem Formulation}
Given a graph $\mathcal{G}$ with normalized adjacency matrix $\matr{A}_{n}\in\mathbb{R}^{N\times N}$ and graph signal (attributes) $\matr{F}\in\mathbb{R}^{N\times P}$, the goal is to find the best partition, i.e., $K$ non-overlapping clusters, $\boldsymbol{\mathcal{C}}=\{\mathcal{C}_1, \mathcal{C}_2,\ldots,\mathcal{C}_K\}$, and the optimal graph filter $\mathcal{H}(\boldsymbol{\Lambda};\matr{\beta})$ with parameters $\matr{\beta}$.   
We  quantify the quality of the clustering based on the filtered graph attributes,  $\matr{\tilde{F}}=\matr{U}\mathcal{H}(\boldsymbol{\Lambda})\matr{U}^\top\matr{F}$, as follows: 
\begin{equation}
\begin{split}
\mathcal{L}(\boldsymbol{\mathcal{C}}, &\mathcal{H}(\boldsymbol{\Lambda};\matr{\beta}))= \sum_{k=1}^{K}\frac{1}{\text{vol}(\mathcal{C}_k)}\sum_{i,j\in \mathcal{C}_k}||\tilde{F}_{i\cdot}-\tilde{F}_{j\cdot}||^2\\
&-\gamma\sum_{k=1}^{K}\frac{1}{\text{vol}(\mathcal{C}_k)}\sum_{\substack{i\in \mathcal{C}_k\\
     j\notin \mathcal{C}_k}}||\tilde{F}_{i\cdot}-\tilde{F}_{j\cdot}||^2,
\end{split}
\label{eq:optprob}
\end{equation}
\noindent where the first and second terms quantify the similarity of the filtered node attributes within clusters 
and between clusters, 
respectively. Thus, we want to minimize the dissimilarity within clusters, i.e., association, while maximizing the separation between clusters, i.e., cut. 
Defining the dissimilarity matrix based on  $\matr{\tilde{F}}$  as $\tilde{W}_{ij}=||\tilde{F}_{i\cdot}-\tilde{F}_{j\cdot}||^2$,  \eqref{eq:optprob} can be rewritten as 

\begin{equation}
\begin{split}
\mathcal{L}(\boldsymbol{\mathcal{C}}, &\mathcal{H}(\boldsymbol{\Lambda};\matr{\beta}))= \sum_{k=1}^{K}\frac{1}{\text{vol}(\mathcal{C}_k)}\sum_{i,j\in \mathcal{C}_k} \tilde{W}_{ij}\\
&-\gamma\sum_{k=1}^{K}\frac{1}{\text{vol}(\mathcal{C}_k)}\sum_{\substack{i\in \mathcal{C}_k\\
     j\notin \mathcal{C}_k}} \tilde{W}_{ij}.
\end{split}
\label{eq:optprobW}
\end{equation}

Our goal is to minimize this cost function in terms of both the graph partition, $\boldsymbol{\mathcal{C}}$ and the graph filter parameters $\beta$. The corresponding  optimization problem can be formulated as
\begin{equation}
\begin{split}
\underset{\boldsymbol{\mathcal{C}},\beta}{\minimize} \hspace{0.1cm} \mathcal{L}(\boldsymbol{\mathcal{C}}, \mathcal{H}(\boldsymbol{\Lambda};\matr{\beta})) + \alpha\mathcal{R}(\boldsymbol{\mathcal{C}}),
\end{split}
\label{eq:minprob}
\end{equation}
\noindent where the regularization term $\mathcal{R}(\boldsymbol{\mathcal{C}})$ will be specified to put additional constraints on the partition such that the connectivity information is also taken into account. We propose a two-step alternating minimization approach to solve this problem, where at each iteration $l$ we first learn the optimal graph partitioning for the given graph signal while the second step learns the optimal graph filter parameters:

\begin{equation}
\begin{split}
&\boldsymbol{\mathcal{C}}^{(l+1)}:=\underset{\boldsymbol{\mathcal{C}}}{\argmin} \hspace{0.1cm} \mathcal{L}(\boldsymbol{\mathcal{C}}, \mathcal{H}(\boldsymbol{\Lambda};\matr{\beta}^{(l)})) + \alpha\mathcal{R}(\boldsymbol{\mathcal{C}}),\\
&\beta^{(l+1)}:=\underset{\beta}{\argmin} \hspace{0.1cm} \mathcal{L}(\boldsymbol{\mathcal{C}}^{(l+1)}, \mathcal{H}(\boldsymbol{\Lambda};\matr{\beta})).
\end{split}
\label{eq:optsteps}
\end{equation}


\subsection{$\boldsymbol{\mathcal{C}}$ update: Clustering}
For the clustering task, given the filtered attributes, $\matr{\tilde{F}}$, we aim to find the graph partition, $\boldsymbol{\mathcal{C}}$. 
Fixing $\mathcal{H}(\boldsymbol{\Lambda};\matr{\beta})$, the cost function in \eqref{eq:optprobW} is closely related to graph based data clustering described in Section \ref{sec:datacluster}. Following the definitions of the normalized cut and association in Eqs. \eqref{eq:cuts} and \eqref{eq:assoc}, respectively, we can rewrite \eqref{eq:optprobW} as 
\begin{equation}
\begin{split}
    \mathcal{L}(\boldsymbol{\mathcal{C}}, \mathcal{H}(\boldsymbol{\Lambda};\matr{\beta}))=\sum_{k=1}^{K}&\frac{\text{links}(\mathcal{C}_k,\mathcal{C}_k)}{\text{vol}(\mathcal{C}_k)}-\gamma\sum_{k=1}^{K}\frac{\text{links}(\mathcal{C}_k,V\setminus \mathcal{C}_k)}{\text{vol}(\mathcal{C}_k)}.\\
\end{split}
\label{eq:links}
\end{equation}

Since it can be shown that $\text{links}(\mathcal{C}_k,V\setminus \mathcal{C}_k)=\text{vol}(\mathcal{C}_k)-\text{links}(\mathcal{C}_k,\mathcal{C}_k)$, the normalized cut is equivalent to the normalized association  \cite{dhillon2004unified}. Therefore, we can rewrite \eqref{eq:links} in terms of the normalized association as 
\begin{equation}
\begin{split}
\mathcal{L}(\boldsymbol{\mathcal{C}}, \mathcal{H}(\boldsymbol{\Lambda};\matr{\beta}))=(1+\gamma)\sum_{k=1}^{K}&\frac{\text{links}(\mathcal{C}_k,\mathcal{C}_k)}{\text{vol}(\mathcal{C}_k)} + K.
\end{split}
\label{eq:links1}
\end{equation}

As mentioned in Section \ref{sec:datacluster}, the problem of minimizing/maximizing normalized association can be rewritten as a trace optimization problem as in Eq. \eqref{eq:traceNassoc}. Therefore, minimizing the cost function in Eq. \eqref{eq:links1} is equivalent to minimizing the following 
\begin{equation}
    \mathcal{L}(\matr{\bar{Z}}, \mathcal{H}(\boldsymbol{\Lambda};\matr{\beta}))=\text{tr}(\matr{\bar{Z}}^\top \matr{D}^{-1/2}\matr{\tilde{W}}\matr{D}^{-1/2}\matr{\bar{Z}}),
    \label{eq:trace}
\end{equation}

\noindent subject to $\matr{\bar{Z}}^\top\matr{\bar{Z}}=\matr{I}$.

\noindent {\em Regularization $\mathcal{R}$:} In order to incorporate the connectivity information into the clustering problem, we propose to use the regularization term $\mathcal{R}(\matr{\bar{Z}})=-\text{tr}(\matr{\bar{Z}}^\top \matr{D}^{-1/2}\matr{A}\matr{D}^{-1/2}\matr{\bar{Z}})$, such that minimizing our regularization term is equivalent to maximizing NAssoc in Eq. \eqref{eq:traceNassocA}. Thus, $\boldsymbol{\mathcal{C}}$ update in Eq. \eqref{eq:optsteps} can be equivalently expressed by the following $\matr{\bar{Z}}$ update equation:

\begin{equation}
\begin{split}
\boldsymbol{\matr{\bar{Z}}}^{(l+1)}
:=\underset{\matr{\bar{Z}},\matr{\bar{Z}}^\top\matr{\bar{Z}}=\matr{I}}{\argmin}& \hspace{0.1cm}  \text{tr}(\matr{\bar{Z}}^\top \matr{D}^{-1/2}\matr{\tilde{W}}\matr{D}^{-1/2}\matr{\bar{Z}})\\
 &-\alpha\text{tr}(\matr{\bar{Z}}^\top \matr{D}^{-1/2}\matr{A}\matr{D}^{-1/2}\matr{\bar{Z}}),\\
 :=\underset{\matr{\bar{Z}},\matr{\bar{Z}}^\top\matr{\bar{Z}}=\matr{I}}{\argmin}& \hspace{0.1cm} \text{tr}(\matr{\bar{Z}}^\top(\matr{\tilde{W}}_n-\alpha\matr{A}_{n})\matr{\bar{Z}}).
\end{split}
\label{eq:probstep1}
\end{equation}



The optimal solution to this problem is the set of eigenvectors corresponding to the $K$ smallest eigenvalues of $\matr{\tilde{W}}_{n}-\alpha\matr{A}_{n}$. The graph partition, $\boldsymbol{\mathcal{C}}^{(l+1)}$, is then updated by applying k-means to the rows of $\matr{\bar{Z}}$.

\vspace{-0.1in}
\subsection{$\beta$ Update: Optimal Filter Design}

Once we have the cluster assignments, we want to determine the coefficients of the optimal filter $\mathcal{H}(\matr{\Lambda};\beta)$. In this paper, we present the derivations for both FIR and IIR filter types, with $\beta=\matr{h}$ for FIR and $\beta=\{\matr{a},\matr{b}\}$ for the ARMA filter.

\subsubsection{FIR Filter}
In this section, we present the optimization problem for learning the parameters  of the optimal polynomial filter with  a given filter order $T$, $\mathcal{H}(\matr{\Lambda})=\sum_{t=0}^{T-1}h_t\matr{\Lambda}^t$, for the clustering task. 

Following the definitions in \cite{segarra2017optimal}, we can define the $t$-th shifted input signal, $\matr{S}^{(t)}\in\mathbb{R}^{N\times P}$, as $\matr{S}^{(t)} := \matr{U}\Lambda^t\matr{U}^\top\matr{F}$ and $\matr{S}_{(i)}$ can then be defined as a $T\times P$ matrix corresponding to the $i$-th node where each row corresponds to the $t$-th shifted input signal at that node with $[\matr{S}_{(i)}]_t := [\matr{S}^{(t)}]_{(i)}$.
With $\matr{\tilde{F}}$ denoting the output of a graph filter for the input signal $\matr{F}$, it follows that
\begin{equation}
    \matr{\tilde{F}} = \matr{U}\left(\sum_{t=0}^{T-1} h_t 
\Lambda^t\right)\matr{U}^\top\matr{F}=\sum_{t=0}^{T-1}h_t \matr{S}^{(t)}.
\end{equation}

Hence, the filtered graph signal corresponding to the $i$-th node 
can be computed as $\matr{\tilde{F}}_i =\sum_{t=0}^{T-1}h_t [\matr{S}^{(t)}]_i= \matr{h}^\top \matr{S}_{(i)}$, with $\matr{h}=[h_0, h_1, \cdots, h_{T-1}]$. 
The cost function in \eqref{eq:optprob} can then be rewritten in terms of the filter coefficient vector, $\matr{h}$, as follows
\begin{equation}
\begin{split}
 \mathcal{L}(\boldsymbol{\mathcal{C}}&, \matr{h})=\sum_{k=1}^{K}\frac{1}{\text{vol}(\mathcal{C}_k)}\sum_{i,j\in \mathcal{C}_k}||\matr{h}^\top \matr{S}_{(i)}-\matr{h}^\top \matr{S}_{(j)}||^2
  \\&-\gamma\sum_{k=1}^{K}\frac{1}{\text{vol}(\mathcal{C}_k)}\sum_{\substack{i\in \mathcal{C}_k\\
     j\notin \mathcal{C}_k}}||\matr{h}^\top \matr{S}_{(i)}-\matr{h}^\top \matr{S}_{(j)}||^2,\\
= \sum_{k=1}^{K}&\frac{1}{\text{vol}(\mathcal{C}_k)}\sum_{i,j\in \mathcal{C}_k}\matr{h}^\top(\matr{S}_{(i)}-\matr{S}_{(j)})(\matr{S}_{(i)}-\matr{S}_{(j)})^\top \matr{h}\\-\gamma\sum_{k=1}^{K}&\frac{1}{\text{vol}(\mathcal{C}_k)}\sum_{\substack{i\in \mathcal{C}_k\\
     j\notin \mathcal{C}_k}}\matr{h}^\top(\matr{S}_{(i)}-\matr{S}_{(j)})(\matr{S}_{(i)}-\matr{S}_{(j)})^\top \matr{h}.
\end{split}
\label{eq:probS}
\end{equation}

Eq. \eqref{eq:probS} can be rewritten as follows:
\begin{equation}
    \mathcal{L}(\boldsymbol{\mathcal{C}}, \matr{h})=(\matr{h}^\top(\matr{B}_T-\gamma \matr{C}_T)\matr{h}),
    \label{eq:probstep2}
\end{equation}

\noindent where $\matr{B}_T$ and $\matr{C}_T$ are $T\times T$ matrices defined as

\begin{equation}
    \matr{B}=\sum_{k=1}^{K}\frac{1}{\text{vol}(\mathcal{C}_k)}\sum_{i,j\in \mathcal{C}_k}(\matr{S}_{(i)}-\matr{S}_{(j)})(\matr{S}_{(i)}-\matr{S}_{(j)})^\top,
\label{eq:B}
\end{equation}

\begin{equation}
    \matr{C}=\sum_{k=1}^{K}\frac{1}{\text{vol}(\mathcal{C}_k)}\sum_{\substack{i\in \mathcal{C}_k\\
     j\notin \mathcal{C}_k}}(\matr{S}_{(i)}-\matr{S}_{(j)})(\matr{S}_{(i)}-\matr{S}_{(j)})^\top.
\label{eq:C}
\end{equation}

Our optimization problem becomes
\begin{equation}
    \underset{\matr{h}}{\minimize}\hspace{0.2cm} (\matr{h}^\top(\matr{B}_T-\gamma \matr{C}_T)\matr{h}).
    \label{eq:probsteph}
\end{equation}

The solution to the optimization problem is the eigenvector of $\matr{B}_T-\gamma\matr{C}_T$ corresponding to the smallest eigenvalue. Once $\matr{h}$ is obtained, the filtered signal can be updated as $\matr{\tilde{F}}=\matr{U}\sum_{t=0}^{T-1}h_t\matr{\Lambda}^t\matr{U}^\top\matr{F}$.

\vspace{0.1in}
\subsubsection{ARMA Filter}

In this section, we present the optimization problem for learning the parameters of an ARMA filter with the following  graph frequency
response
\begin{equation}
    \mathcal{H}(\boldsymbol{\Lambda})=\frac{\sum_{t=0}^{T-1} a_t\boldsymbol{\Lambda}^t}{\matr{I}+\sum_{q=1}^{Q-1} b_q\boldsymbol{\Lambda}^q},
    \label{eq:armafilter}
\end{equation}

\noindent where $\matr{a}=[a_0, a_1, \dots, a_{T-1}]$ and  $\matr{b}=[b_1, b_2, \dots, b_{Q-1}]$ are the filter coefficients, and $(T,Q)$ is the pair of filter orders. 

In order to find the filter parameters $\matr{a}$ and $\matr{b}$ we introduce an auxiliary polynomial $\Gamma_{M}(\lambda)=\sum_{m=0}^{M-1} c_m\lambda^m$. $\Gamma_{M}(\lambda)$ can be viewed as the reciprocal polynomial of $(1+\sum_{q=1}^{Q-1} b_q\lambda^q)$. In general, the reciprocal polynomial has a larger order than the denominator polynomial, i.e., $M\geq Q$ \cite{xu2023arma}. Letting $\sum_{m=0}^{M-1} c_m\boldsymbol{\Lambda}^m=\frac{1}{\matr{I}+\sum_{q=1}^{Q-1} b_q\boldsymbol{\Lambda}^q}$, Eq. \eqref{eq:armafilter} becomes 

\begin{equation}
    \mathcal{H}(\boldsymbol{\Lambda})=\sum_{m=0}^{M-1} c_m\boldsymbol{\Lambda}^m\sum_{t=0}^{T-1} a_t\boldsymbol{\Lambda}^t.
    \label{eq:armafilter1}
\end{equation}

To solve for the filter parameters, we introduce a constraint to assure that $(\sum_{m=0}^{M-1} c_m\boldsymbol{\Lambda}^m)(\matr{I}+\sum_{q=1}^{Q-1} b_q\boldsymbol{\Lambda}^q)=\matr{I}$.  
We can rewrite the polynomials as $\sum_{m=0}^{M-1} c_m\boldsymbol{\Lambda}^m= \text{diag}(\boldsymbol{\Psi}_{M}\matr{c})$ and $\sum_{q=1}^{Q-1} b_q\boldsymbol{\Lambda}^q= \text{diag}(\boldsymbol{\bar{\Psi}}_{Q}\matr{b})$, where $\boldsymbol{\Psi}_{M}$ and $\boldsymbol{\bar{\Psi}}_{Q}$ are defined as
\begin{equation*}
\boldsymbol{\Psi}_{M}=
    \begin{bmatrix}
1& \lambda_1 & \lambda_1^2 & \cdots & \lambda_1^{M-1}\\
1& \lambda_2 & \lambda_2^2 & \cdots & \lambda_2^{M-1}\\
\vdots &\vdots & \vdots & \ddots & \vdots\\
1& \lambda_N & \lambda_N^2 & \cdots & \lambda_N^{M-1}\\
    \end{bmatrix},\\
\end{equation*}
\begin{equation*}
    \boldsymbol{\bar{\Psi}}_{Q}=
    \begin{bmatrix}
\lambda_1 & \lambda_1^2 & \cdots & \lambda_1^{Q-1}\\
\lambda_2 & \lambda_2^2 & \cdots & \lambda_2^{Q-1}\\
\vdots & \vdots & \ddots & \vdots\\
\lambda_N & \lambda_N^2 & \cdots & \lambda_N^{Q-1}\\
    \end{bmatrix}.
\end{equation*}

\noindent Using these definitions we can then rewrite $(\sum_{m=0}^{M-1} c_m\boldsymbol{\Lambda}^m)(\matr{I}+\sum_{q=1}^{Q-1} b_q\boldsymbol{\Lambda}^q)=\matr{I}$ as $(\text{diag}(\boldsymbol{\Psi}_{M}\matr{c}))(\matr{I}+\text{diag}(\boldsymbol{\bar{\Psi}}_{Q}\matr{b}))=\matr{I}$ and the optimization problem becomes
\begin{equation}
\begin{split}
&\underset{\matr{a},\matr{b}, \matr{c}}{\minimize} \hspace{0.1in} \mathcal{L}(\boldsymbol{\mathcal{C}}, \mathcal{H}(\boldsymbol{\Lambda};\matr{[a,c]}))\\
&+||\text{diag}(\boldsymbol{\Psi}_M\matr{c})[ \matr{I}+\text{diag}(\boldsymbol{\bar{\Psi}}_Q\matr{b})]-\matr{I}||_F^2.
\end{split}
\label{eq:minprob2}
\end{equation}





In order to find the parameters $\matr{a}$, $\matr{b}$ and $\matr{c}$, we propose an alternating minimization approach where in order to learn each of the variables we fix the other two. 
Thus, the $\beta$ update in \eqref{eq:optsteps} becomes

\begin{equation}
\begin{split}
\boldsymbol{\matr{a}}^{(l+1)}:=&\underset{\matr{a}}{\argmin} \hspace{0.1cm} \mathcal{L}(\boldsymbol{\mathcal{C}}, \mathcal{H}(\boldsymbol{\Lambda};[\matr{a},\matr{c}^{(l)}])),\\
\boldsymbol{\matr{b}}^{(l+1)}:=&\underset{\matr{b}}{\argmin} \hspace{0.1cm} 
||\text{diag}(\boldsymbol{\Psi}_M\matr{c}^{(l)})[ \matr{I}+\text{diag}(\boldsymbol{\bar{\Psi}}_Q\matr{b})]-\matr{I}||_F^2,\\
\boldsymbol{\matr{c}}^{(l+1)}:=&\underset{\matr{c}}{\argmin} \hspace{0.1cm} \mathcal{L}(\boldsymbol{\mathcal{C}}, \mathcal{H}(\boldsymbol{\Lambda};[\matr{a}^{(l+1)},\matr{c}])) \\
+&||\text{diag}(\boldsymbol{\Psi}_M\matr{c})[ \matr{I}+\text{diag}(\boldsymbol{\bar{\Psi}}_Q\matr{b}^{(l+1)})]-\matr{I}||_F^2.\\
\end{split}
\label{eq:optsteps2}
\end{equation}


\noindent {\em \textbf{Update a:}} In order to update $\matr{a}$, we define the $t$-th shifted input signal as $\matr{S}^{(t)} := \matr{U}(\sum_{m=0}^{M-1} c_m\boldsymbol{\Lambda}^m)\boldsymbol{\Lambda}^t\matr{U}^\top\matr{F}$ and $\matr{S}_{(i)}$ as a $T\times P$ matrix corresponding to the $i$-th node where each row corresponds to the $t$-th shifted input signal at that node, $[\matr{S}^{(t)}]_{(i)}$. Therefore, $\matr{\tilde{F}}_i =\sum_{t=0}^{T-1}a_t [\matr{S}^{(t)}]_i= \matr{a}^\top \matr{S}_{(i)}$. 

The cost function in \eqref{eq:optprob} can be rewritten in terms of the filter coefficients $\matr{a}$ and the newly defined $\matr{S}_{(i)}$ as follows
\begin{equation}
\begin{split}
 \mathcal{L}(\boldsymbol{\mathcal{C}}, &\matr{a},\matr{c})= \sum_{k=1}^{K}\frac{1}{\text{vol}(\mathcal{C}_k)}\sum_{i,j\in \mathcal{C}_k}||\matr{a}^\top \matr{S}_{(i)}-\matr{a}^\top \matr{S}_{(j)}||^2
 \\&-\gamma\sum_{k=1}^{K}\frac{1}{\text{vol}(\mathcal{C}_k)}\sum_{\substack{i\in \mathcal{C}_k\\
     j\notin \mathcal{C}_k}}||\matr{a}^\top \matr{S}_{(i)}-\matr{a}^\top \matr{S}_{(j)}||^2,\\
= \sum_{k=1}^{K}&\frac{1}{\text{vol}(\mathcal{C}_k)}\sum_{i,j\in \mathcal{C}_k}\matr{a}^\top(\matr{S}_{(i)}-\matr{S}_{(j)})(\matr{S}_{(i)}-\matr{S}_{(j)})^\top \matr{a}\\-\gamma\sum_{k=1}^{K}&\frac{1}{\text{vol}(\mathcal{C}_k)}\sum_{\substack{i\in \mathcal{C}_k\\
     j\notin \mathcal{C}_k}}\matr{a}^\top(\matr{S}_{(i)}-\matr{S}_{(j)})(\matr{S}_{(i)}-\matr{S}_{(j)})^\top \matr{a}.
\end{split}
\label{eq:probSb}
\end{equation}

Eq. \eqref{eq:probSb} can be rewritten as follows:
\begin{equation}
    \mathcal{L}(\boldsymbol{\mathcal{C}}, \matr{a}, \matr{c})=(\matr{a}^\top(\matr{B}_T-\gamma \matr{C}_T)\matr{a}),
    \label{eq:probstep2}
\end{equation} 
\noindent where $\matr{B}_T$ and $\matr{C}_T$ are $T\times T$ matrices defined as in Eqs. \eqref{eq:B} and \eqref{eq:C} using the newly defined $\matr{S}_{(i)}$.
And, our optimization problem becomes
\begin{equation}
    \underset{\matr{a}}{\minimize}\hspace{0.2cm} (\matr{a}^\top(\matr{B}_T-\gamma \matr{C}_T)\matr{a}).
    \label{eq:probstep2}
\end{equation}

\noindent whose solution is the eigenvector of $\matr{B}_T-\gamma\matr{C}_T$ corresponding to the smallest eigenvalue.



\vspace{0.1in}
\noindent {\em \textbf{Update b:}} Vectorizing each term in \eqref{eq:optsteps2} for the $\matr{b}$ update, we have the following equivalent optimization problem
\begin{equation}
    \underset{\matr{b}}{\minimize}\hspace{0.2cm} ||\boldsymbol{\Psi}_M\matr{c}+ \text{diag}(\boldsymbol{\Psi}_M\matr{c})\boldsymbol{\bar{\Psi}}_Q\matr{b}-\matr{1}_N||^2,
\end{equation}
\noindent where $\matr{1}_N\in\mathbf{R}^N$ is an all ones vector. To find $\matr{b}=[b_1, b_2, \cdots, b_{Q-1}]$, the following objective function $\mathcal{L}_1(\matr{b})=||\boldsymbol{\Psi}_M\matr{c}+ \text{diag}(\boldsymbol{\Psi}_M\matr{c})\boldsymbol{\bar{\Psi}}_Q\matr{b}-\matr{1}_N||^2$
is minimized with respect to $\matr{b}$ by setting
\begin{equation*}
\begin{split}  
    \nabla_b \mathcal{L}_1=2\boldsymbol{\bar{\Psi}}_Q^\top\text{diag}&(\boldsymbol{\Psi}_M\matr{c})[\boldsymbol{\Psi}_M\matr{c}+ \text{diag}(\boldsymbol{\Psi}_M\matr{c})\boldsymbol{\bar{\Psi}}_Q\matr{b}]\\
    &-2\boldsymbol{\bar{\Psi}}_Q^\top\text{diag}(\boldsymbol{\Psi}_M\matr{c})\matr{1}_N=0.
\end{split}
\end{equation*}
\\
After some algebraic manipulations, $\matr{b}$ can be updated as
\begin{equation}
\begin{split}  
\matr{b}=\matr{Y}_1^{-1}\matr{v}_1,
\end{split}
\label{eq:solveb}
\end{equation}

\noindent where $\matr{Y}_1\in\mathbb{R}^{Q-1\times Q-1}$ and $\matr{v}_1\in\mathbb{R}^{Q-1}$ are defined as $\matr{Y}_1=(\boldsymbol{\bar{\Psi}}_Q^\top\text{diag}(\boldsymbol{\Psi}_M\matr{c})\text{diag}(\boldsymbol{\Psi}_M\matr{c})\boldsymbol{\bar{\Psi}}_Q)$ and $\matr{v}_1=\boldsymbol{\bar{\Psi}}_Q^\top\text{diag}(\boldsymbol{\Psi}_M\matr{c})[\matr{1}_N-\boldsymbol{\Psi}_M\matr{c}]$, respectively. $\matr{Y}_1$ is full rank and thus $\matr{Y}_1^{-1}$ exists.

\vspace{0.1in}

\noindent {\em \textbf{Update c:}} We can define the $m$-th shifted input signal as $\matr{S}^{(m)} := \matr{U}\boldsymbol{\Lambda}^m(\sum_{t=0}^{T-1} a_t\boldsymbol{\Lambda}^t)\matr{U}^\top\matr{F}$ and $\matr{\tilde{F}}_i =\sum_{m=0}^{M-1}c_m [\matr{S}^{(m)}]_i= \matr{c}^\top \matr{S}_{(i)}$, with $\matr{S}_{(i)}$ being a $M\times P$ matrix corresponding to the $i$-th node where each row corresponds to the $m$-th shifted input signal at that node, $[\matr{S}^{(m)}]_{(i)}$.

The $\matr{c}$ update in \eqref{eq:optsteps2} can be rewritten as 
\begin{equation}
\begin{split}
    \underset{\matr{c}}{\minimize}\hspace{0.1cm} &\matr{c}^\top(\matr{B}_M-\gamma \matr{C}_M)\matr{c} \\
    &+ ||\boldsymbol{\Psi}_M\matr{c}+ \text{diag}(\boldsymbol{\bar{\Psi}}_Q\matr{b})\boldsymbol{\Psi}_M\matr{c}-\matr{1}_N||^2.
\end{split}
\end{equation}

To find $\matr{c}=[c_0, c_1, \cdots, c_{M-1}]$, the following objective function $\mathcal{L}_2(\matr{c})=\matr{c}^\top(\matr{B}_M-\gamma \matr{C}_M)\matr{c}+||\boldsymbol{\Psi}_M\matr{c}+ \text{diag}(\boldsymbol{\bar{\Psi}}_Q\matr{b})\boldsymbol{\Psi}_M\matr{c}-\matr{1}_N||^2$
is minimized with respect to $\matr{c}$ by setting
\begin{equation*}
\begin{split}  
   \nabla_c \mathcal{L}_2&=2(\matr{B}_M-\gamma \matr{C}_M)\matr{c}\\
   & +2\boldsymbol{\Psi}_M^\top(\matr{I}+\text{diag}(\boldsymbol{\bar{\Psi}}_Q\matr{b}))^\top(\matr{I}+\text{diag}(\boldsymbol{\bar{\Psi}}_Q\matr{b}))\boldsymbol{\Psi}_M\matr{c}\\
   &-2\boldsymbol{\Psi}_M^\top(\matr{I}+\text{diag}(\boldsymbol{\bar{\Psi}}_Q\matr{b})^\top)\matr{1}_N=0.
\end{split}
\end{equation*}

After some algebraic manipulations, $\matr{c}$ can be updated as
\begin{equation}
\begin{split}  
\matr{c}=\matr{Y}_2^{-1}\matr{v}_2,
\end{split}
\label{eq:solvec}
\end{equation}

\noindent where $\matr{Y}_2\in\mathbb{R}^{M\times M}$ and $\matr{v}_2\in\mathbb{R}^{M}$ are defined as $\matr{Y}_2=(\matr{B}_M-\gamma \matr{C}_M)+\boldsymbol{\Psi}_M^\top(\matr{I}+\text{diag}(\boldsymbol{\bar{\Psi}}_Q\matr{b}))^\top(\matr{I}+\text{diag}(\boldsymbol{\bar{\Psi}}_Q\matr{b}))\boldsymbol{\Psi}_M$ and $\matr{v}_2=\boldsymbol{\Psi}_M^\top(\matr{I}+\text{diag}(\boldsymbol{\bar{\Psi}}_Q\matr{b})^\top)\matr{1}_N$, respectively. $\matr{Y}_2$ is full rank and thus $\matr{Y}_2^{-1}$ exists.

\vspace{0.1in}

We solve iteratively for the filter coefficients  $\matr{a}$, $\matr{b}$, and $\matr{c}$ until convergence. Once the filter coefficients are obtained at the $l$-th iteration, we update the filtered signal $\matr{\tilde{F}}^{(l)}$. 
Both variable updates, $\boldsymbol{\mathcal{C}}$ and $\beta$, for Clustering and Optimal Filter Design steps, respectively, are repeated until convergence as described in Algorithm \ref{alg:gf}.

\begin{algorithm}[h]
\caption{GraFiCA}
\label{alg:gf}
\begin{algorithmic}[1]
\newcommand{\IndState}[1][1]{\STATE\hspace{0.2in}}
\renewcommand{\algorithmicrequire}{\textbf{Input:}}
\Require Normalized adjacency matrix $\matr{A}_{n}$, graph signal $\matr{F}$, number of clusters $K$, parameters $\alpha, \gamma$, filter orders $(T,Q)$, and $M\geq Q.$
\renewcommand{\algorithmicrequire}{\textbf{Output:}} 
\Require Cluster partition $\boldsymbol{\mathcal{C}}$, graph filter $\mathcal{H}(\matr{L})$.
\State $\matr{L}_{n}= \matr{U}\boldsymbol{\Lambda}\matr{U}^\top$
\State [NMI$^{(0)}$, $\boldsymbol{\mathcal{C}}^{(0)}$]=ClusteringStep($\matr{F}$) 
\State Initialize $\matr{c}^{(0)}$ for ARMA
\State $l=0$
\While{$|\text{NMI}^{(l)}-\text{NMI}^{(l-1)}|>10^{-3}$}
\If{$Q=1$} \Comment{FIR Filter}
\State Compute $\matr{B}_T$ and $\matr{C}_T$ using \eqref{eq:B} and \eqref{eq:C}
\State $\matr{S}\leftarrow(\matr{B}_T-\gamma\matr{C}_T)$
\State $\matr{S}= \matr{H}\Gamma\matr{H}^\top$
\State $\matr{h}\leftarrow \matr{H}_{\cdot 1}$
\State $\matr{\tilde{F}}\leftarrow\matr{U}\sum_{t=0}^{T-1}h_t\matr{\Lambda}^t\matr{U}^\top\matr{F}$ 
\Else \Comment{ARMA Filter}
\State $r=0$
\While{$||\matr{a}^{(r)}-\matr{a}^{(r-1)}||^2>10^{-3}$, $||\matr{b}^{(r)}-\matr{b}^{(r-1)}||^2>10^{-3}$, and $||\matr{c}^{(r)}-\matr{c}^{(r-1)}||^2>10^{-3}$}
\State Compute $\matr{B}_T$ and $\matr{C}_T$ using \eqref{eq:B} and \eqref{eq:C}
\State $\matr{S}\leftarrow(\matr{B}_T-\gamma\matr{C}_T)$
\State $\matr{S}= \matr{H}\Gamma\matr{H}^\top$
\State $\matr{a}^{(r)}\leftarrow \matr{H}_{\cdot 1}$
\State $\matr{b}^{(r)}\leftarrow\matr{Y}_1^{-1}\matr{v}_1$ using \eqref{eq:solveb}
\State $\matr{c}^{(r)}\leftarrow\matr{Y}_2^{-1}\matr{v}_2$ using \eqref{eq:solvec}
\State $r=r+1$
\EndWhile
\State $\matr{\tilde{F}}\leftarrow\matr{U}\frac{\sum_{t=0}^{T-1} a_t\boldsymbol{\Lambda}^t}{\matr{I}+\sum_{q=1}^{Q-1} b_q\boldsymbol{\Lambda}^q}\matr{U}^\top\matr{F}$ 

\EndIf
\State  [NMI$^{(l)}$,$\boldsymbol{\mathcal{C}}^{(l)}$]=ClusteringStep($\matr{\tilde{F}}$) 
\State $l=l+1$
\EndWhile
\Function{ClusteringStep}{$\matr{F}$}
  \State $W_{ij}\leftarrow ||F_{i\cdot}-F_{j\cdot}||^2$
\State $\matr{W}'\leftarrow\matr{W}_{n}-2\alpha\matr{A}_{n}$
\State  $\matr{W}'= \matr{V}\Gamma \matr{V}^\top$ 
\State  $\matr{V}\leftarrow \matr{V}(:,1:K)$ 
\State $\boldsymbol{\mathcal{C}}\leftarrow k\text{-means}(\matr{V},K)$
\State Compute NMI 
\State \Return NMI and $\boldsymbol{\mathcal{C}}$
\EndFunction
\end{algorithmic}
\end{algorithm}
\vspace{-0.1in}
\section{Computational Complexity}

The computational complexity of the algorithm is mostly due to the eigendecompositions and inverse operations at each iteration. There is only one full eigendecomposition at the beginning of the algorithm for the normalized Laplacian of the graph. Eigendecompositions of a $N\times N$ matrix, in general, have complexity on the order $\mathcal{O}(N^3)$. However, it is important to note that this is the worst-case scenario, and there exist fast algorithms that will reduce this to $\mathcal{O}(N^e)$, $2<e<2.376$ \cite{demmel2007fast}. Moreover, there are algorithms that approximate the spectral decomposition of graphs \cite{coutino2020fast} reducing the computational complexity to $\mathcal{O}(N^2)$. At each iteration, we find the clusters by
computing the eigenvectors corresponding to the $K$ smallest eigenvalues of $\matr{\tilde{W}}_{n}-2\alpha\matr{A}_{n}$. The computational complexity of finding the $K$ eigenvectors corresponding the smallest eigenvalues is  $\mathcal{O}(N^2K)$. The filter coefficients $\matr{h}$ for FIR and $\matr{a}$ for ARMA are found by computing the eigenvector corresponding to the smallest eigenvalue of  $T\times T$  matrices, with computational complexity $\mathcal{O}(T^2)$. For finding $\matr{b}$ and $\matr{c}$, we have an inverse operation, and the standard matrix inversion algorithm has a time complexity of $\mathcal{O}((Q-1)^3)$ and $\mathcal{O}((M)^3)$ for finding $\matr{b}$ and $\matr{c}$, respectively.
It is important to note that $K,T, Q, M<<N$, so the total complexity is dominated by $\mathcal{O}(N^2K)$.
\begin{table}[h!]
    \centering
    \footnotesize
    \begin{tabular}{c|cc}
    \hline
$\matr{L}_n=\matr{U}\boldsymbol{\Lambda}\matr{U}^\top$ & \multicolumn{2}{c}{$\mathcal{O}(N^e)$, $2<e<2.376$}
        \\
        \hline
        Clustering Step & \multicolumn{2}{c}{$\mathcal{O}(N^2K)$}\\
        \hline
        \multirow{4}{*}{Filter Design Step} & FIR & ARMA\\
        & $\matr{h}$: $\mathcal{O}(T^2)$ & $\matr{a}$: $\mathcal{O}(T^2)$ \\
        & & $\matr{b}$: $\mathcal{O}((Q-1)^3)$\\
        & & $\matr{c}$: $\mathcal{O}((M)^3)$ \\
        \hline
        Total & \multicolumn{2}{c}{$\mathcal{O}(N^2K)$}\\
        \hline
        \end{tabular}
    \caption{Computational complexities of GraFiCA steps.}
    \label{tab:complexity}
\end{table}
\vspace{-0.2in}
\section{Results}
\label{sec:experiments}
We evaluate the proposed graph filter learning method for both FIR and ARMA filters on five attributed networks. The first three, Cora, Citeseer, and PubMed \cite{sen2008collective}, are citation networks where the nodes correspond to publications, and the edges correspond to citations among the papers. Cora has 2,708 machine learning papers classified into seven classes: case-based reasoning, genetic algorithms, neural networks, probabilistic methods, reinforcement learning, rule learning, and theory. Citeeser has 3,327 machine learning publications classified into 6 classes: agents, artificial intelligence, database, information retrieval, machine learning, and human-computer interaction. PubMed has 19,717 papers classified into one of three classes: Diabetes Mellitus -Experimental, Diabetes Mellitus type 1, and Diabetes Mellitus type 2. Wiki \cite{yang2015network} is a webpage network where the nodes are webpages and the edges are the links between them. It contains 2,405 long text documents classified into 17 classes. Sinanet \cite{jia2017node} is a microblog users' relationship network where the edges between the users represent the followers/followees relationships. It contains 3,490 users from 10 major forums, including finance and economics, literature and arts, fashion and vogue, current events and politics, sports, science and technology, entertainment, parenting and education, public welfare, and normal life. The nodes in Cora and Citeseer are associated with binary word vectors indicating the presence or absence of some words, and the nodes in PubMed and Wiki are associated with tf-idf weighted word vectors. Sinanet is associated with a tf-idf weighted vector indicating the users' interest distribution in each forum.  Table \ref{tab:datasets} summarizes the details for each dataset.
\vspace{-0.1in}
\begin{table}[h]
\footnotesize
  \centering
  \caption{Datasets statistics} 
    \begin{tabular}{c|c|c|c|c|c}
    \hline
      \centering
      Dataset & Cora & Citeseer & Sinanet & Wiki & PubMed\\
      \hline
      Nodes & 2708 & 3327 & 3490 & 2405 & 19717\\
      Links & 5429 & 4732 & 30282 & 17981 & 44338\\
      Features & 1433 & 3703 & 10 & 4973 & 500\\
      Classes & 7 & 6 & 10 & 17 & 3\\
\hline
    \end{tabular}
    \label{tab:datasets}
    \end{table}
    
\vspace{-0.1in}    
\subsection{Baseline Methods and Metrics}
In order to evaluate the performance of our method and the importance of taking into account both the topology and the attributes of a graph, we compare our method with three classes of methods. 
\begin{itemize}
    \item The first class of methods, $k$-means, spectral clustering, and  CGFKM \cite{du2024k} only use the node attributes.  CGFKM is a k-means based method that learns a Chebyshev polynomial approximation graph filter similar to our graph learning stage. However, this method does not use the connectivity of the network.
    \item The second class of methods only use the graph topology. Spectral clustering on the graph, SC-G, uses eigendecomposition on the graph Laplacian, Louvain \cite{blondel2008fast} is a well-known community detection method, and Multi-Scale Community Detection (MS-CD) \cite{tremblay2014graph}, uses graph spectral wavelets to compute a similarity metric between nodes to find communities at different scales.
    \item Finally, the third class of methods include AGC \cite{zhang2019attributed}, GAE and VGAE \cite{kipf2016variational}, EGAE \cite{zhang2022embedding}, ARGE and ARVGE \cite{pan2018adversarially}, which  use both node attributes and graph structure. AGC is based on spectral graph filtering, GAE and VGAE are benchmark autoencoder-based methods, while ARGE and ARVGE are benchmark adversarial GAE methods. EGAE is a GAE based model designed specifically for graph clustering.
\end{itemize}

The clustering performance of the different methods is quantified using Normalized Mutual Information (NMI) \cite{danon2005comparing} and Adjusted Rand Index (ARI) \cite{hubert1985comparing}. 

\subsection{Performance Evaluation}
The input parameters required by our method are the number of clusters $K$, the filter orders $(T,Q)$ and the hyperparameters, $\alpha$ for the clustering step, and $\gamma$ for filter optimization step.
For selecting the best $\alpha$ and $\gamma$ values, we tested $\alpha,\gamma\in[0,0.5]$, and the results for the filter with the best performance in terms of NMI are reported. 
For the FIR filter, we evaluated different values for the filter order $T$ between 3 and 10 to determine the order that gives the best NMI for each dataset. For ARMA, we assume that $Q>T$  for an order pair $(T,Q)$ \cite{liu2018filter}. 
The optimal values of the parameters $\alpha$, $\gamma$, and the filter orders $T$ for FIR, $(T,Q)$ for ARMA are listed for each dataset in Table \ref{tab:param}. 
In all of the tested datasets, the number of clusters is known. For the baseline methods, the parameter settings reported in the original papers are used \cite{zhang2019attributed,kipf2016variational,pan2018adversarially}. 
\begin{table}[h]
\footnotesize
  \centering
  \caption{The optimal parameters for the different datasets.} 
    \begin{tabular}{c|c|c|c|c|c|c}
    \hline
      \centering
    & Dataset &   Cora & Citeseer & Sinanet & Wiki & PubMed\\ 
    \hline
   \multirow{3}{*}{FIR} &$\alpha$ & 0.056 & 0.06 & 0.044 & 0.001 & 0.01\\
    &$\gamma$ & 0.074 & 0.10 & 0.022 & 0.03 & 0.42\\
   & $T$ &3 &3  &3  &3 &  3\\
   \hline
      \multirow{3}{*}{ARMA} &$\alpha$ & 0.05 & 0.058 & 0.05& 0.001 & 0.01\\
    &$\gamma$ & 0.08 & 0.101 & 0.03 & 0.028 & 0.40 \\
    &$(T,Q)$ & (2,3) & (2,3) & (2,3) & (3,4) & (2,3) \\
    \hline
    \end{tabular}
    \label{tab:param}
    \end{table}

\begin{table*}[h!]
\footnotesize
  \centering
  \caption{Normalized Mutual Information (NMI) and Adjust Rand Index (ARI) results.} 
    \begin{tabular}{l|c|c|c|c|c|c|c|c|c|c}
    \hline
 & \multicolumn{2}{c|}{Cora} & \multicolumn{2}{c|}{Citeseer} & \multicolumn{2}{c|}{Sinanet} & \multicolumn{2}{c|}{Wiki} & \multicolumn{2}{c}{PubMed} \\
 \cline{2-11}
    Algorithms & NMI & ARI & NMI & ARI & NMI & ARI & NMI & ARI & NMI & ARI\\ 
    \hline
    k-means & 
    0.2825 &0.1621 & 0.3597 & 0.3279 & 0.6413 & 0.5828 & 0.3323 & 0.0578 & 0.2053& 0.1810\\
    SC-F    &0.2856 & 0.1615 & 0.3244 & 0.2898 & 0.5395 & 0.3802 & 0.3974 & 0.0954& 0.0164 & 0.0098\\
    CGFKM \cite{du2024k} & 0.1327 & 0.0522 & 0.0773 & 0.0011 & 0.6415 & 0.5664 & 0.0765 & 0.0025 & 0.1347 &0.0911\\
     \hline
     SC-G & 0.0872 & 0.0185 & 0.0573 & 0.0051 & 0.1653& 0.0444 & 0.1709 & 0.0073& 0.0075& 0.0013\\
    Louvain \cite{blondel2008fast} & 0.5044 & 0.3433 & 0.3645 & 0.3457 & 0.2490 & 0.2044 & 0.4105 & 0.2369 & 0.2059 & 0.1099\\
    MS-CD \cite{tremblay2014graph} &  0.5072 & 0.3453 & 0.4064 & 0.3926 & 0.4064 & 0.1997 & 0.3548 & 0.2677 & 0.1773 & 0.1465 \\
    \hline
    AGC \cite{zhang2019attributed} & 0.5170 & 0.3982 & 0.4086 & 0.4124& 0.5573 & 0.3892 & 0.4268 & 0.1440 & 0.3158 & \textbf{0.3105}\\
 GAE \cite{kipf2016variational} &0.4209 & 0.3160 & 0.1706 & 0.1018
& 0.4346 & 0.3523  & 0.1316 & 0.0781  & 0.2374 & 0.1955
 \\
    VGAE \cite{kipf2016variational} & 0.4276 & 0.3188 & 0.2112 & 0.1440 & 0.4567 & 0.3881 & 0.2982 & 0.1143 & 0.2403 & 0.2119
 \\
  EGAE \cite{zhang2022embedding}	& 0.5401 & 0.4723 &	0.4122 & \underline{0.4324}	&  0.3404 & 0.2773  & 0.4711 & \textbf{0.3308}	&0.3205 & 0.2893\\
    ARGE \cite{pan2018adversarially}& 0.4562 & 0.3865 & 0.2967 & 0.2781 & 0.4815 & 0.3781  & 0.3715 & 0.1129  & 0.2359 & 0.2258\\
    ARVGE \cite{pan2018adversarially} & 0.4657 & 0.3895 & 0.3124 & 0.3022 & 0.4854& 0.3993 &0.3987 & 0.1084 & 0.0826 & 0.0373\\
    \hline
 $\text{GraFiCA}_\text{\tiny{FIR}}$ &\textbf{0.5465} & \textbf{0.4743} & \underline{0.4228} & 0.4283 & \textbf{0.6578} & \underline{0.6721} & \underline{0.5125} & 0.2771 & \textbf{0.3279} & \underline{0.2995}  \\
    $\text{GraFiCA}_\text{\tiny{ARMA}}$ & \underline{0.5421} & \underline{0.4746} & \textbf{0.4261} & \textbf{0.4365}& \underline{0.6561} & \textbf{0.6736} & \textbf{0.5150} & \underline{0.2807} & \underline{0.3265} & 0.2915 \\
    \hline
    \end{tabular}%
  \label{tab:results}%
\end{table*}%

\begin{figure*}[h]
    \centering
    \begin{subfigure}[b]{0.19\linewidth}
    \includegraphics[width=\linewidth]{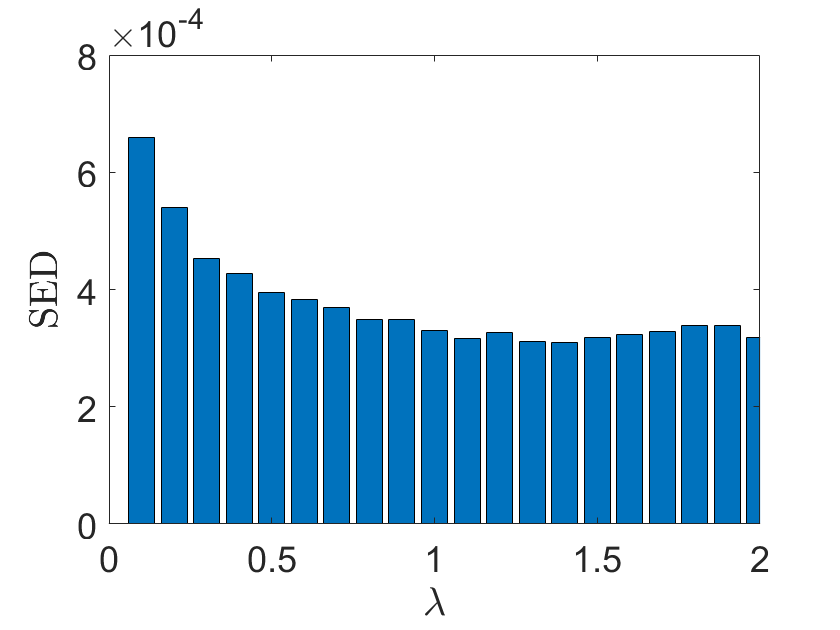}
        \caption{}
    \label{fig:coraSED}
    \end{subfigure}
    \begin{subfigure}[b]{0.19\linewidth}
    \includegraphics[width=\linewidth]{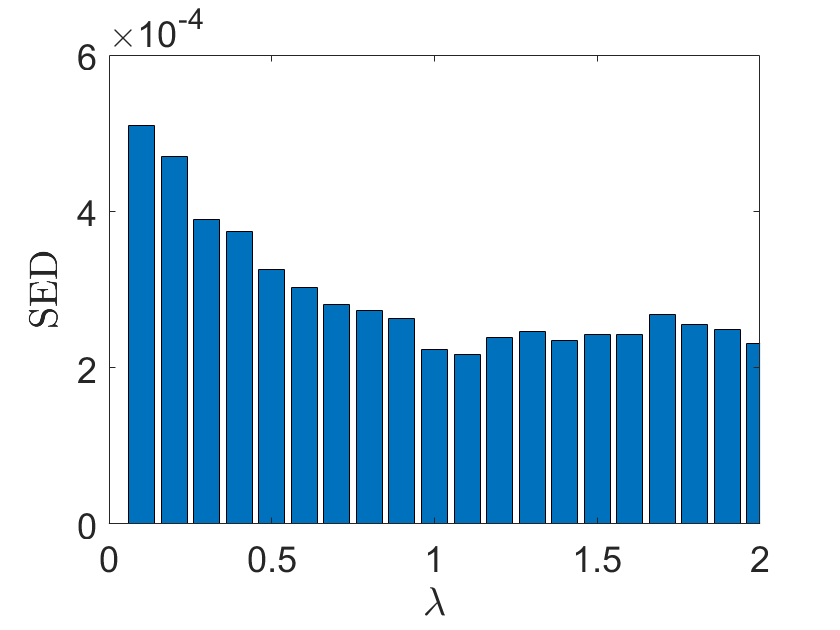}
        \caption{}
    \label{fig:citeSED}
    \end{subfigure}
    \begin{subfigure}[b]{0.19\linewidth}
    \includegraphics[width=\linewidth]{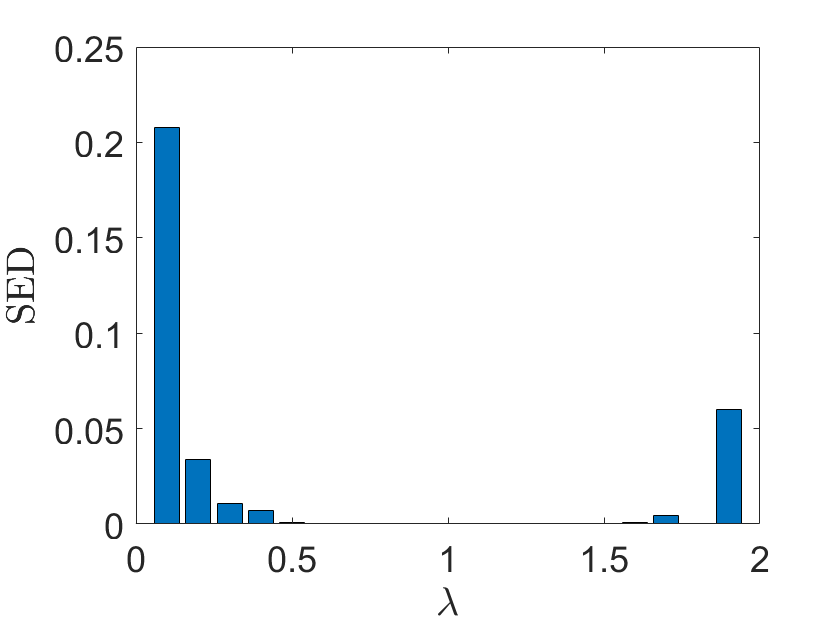}
        \caption{}
    \label{fig:sinaSED}
    \end{subfigure}
    \begin{subfigure}[b]{0.19\linewidth}
    \includegraphics[width=\linewidth]{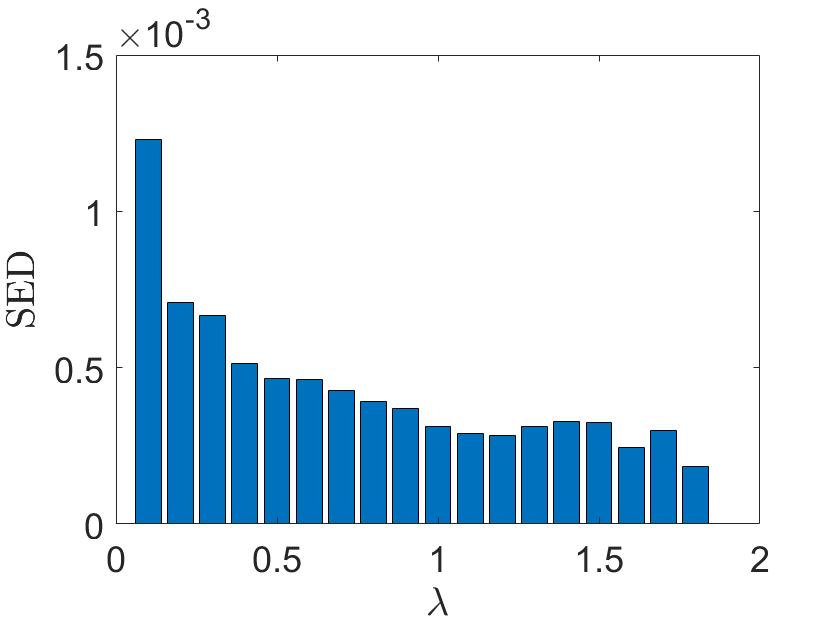}
        \caption{}
    \label{fig:wikiSED}
    \end{subfigure}
    \begin{subfigure}[b]{0.19\linewidth}
    \includegraphics[width=\linewidth]{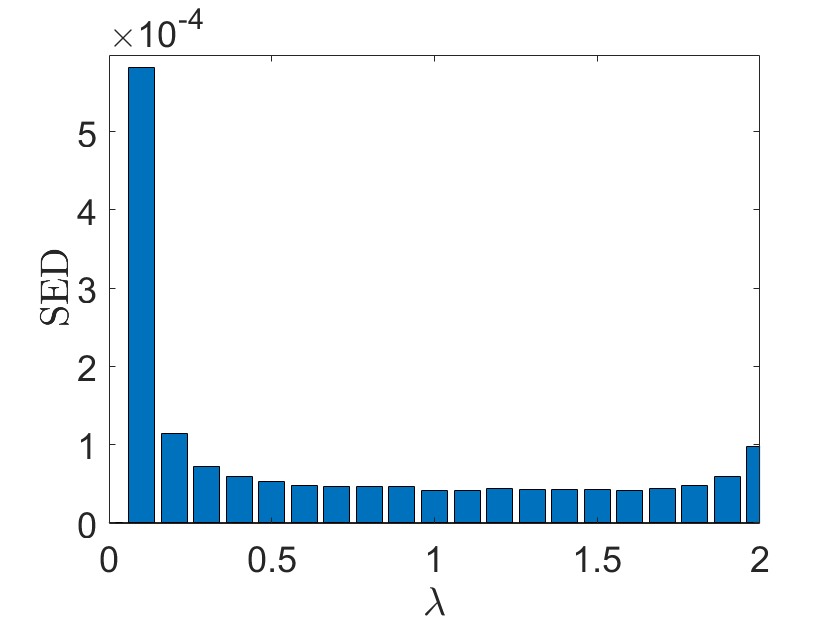}
    \caption{}
    \label{fig:pubSED}
    \end{subfigure}
    \centering
    \begin{subfigure}[b]{0.19\linewidth}
    \includegraphics[width=\linewidth]{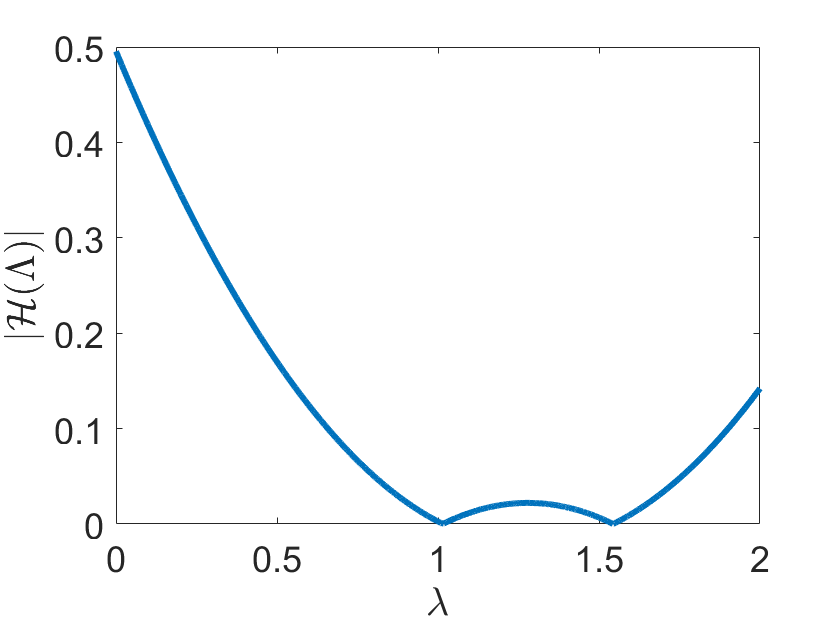}
        \caption{}
    \label{fig:corafilt}
    \end{subfigure}
    \begin{subfigure}[b]{0.19\linewidth}
    \includegraphics[width=\linewidth]{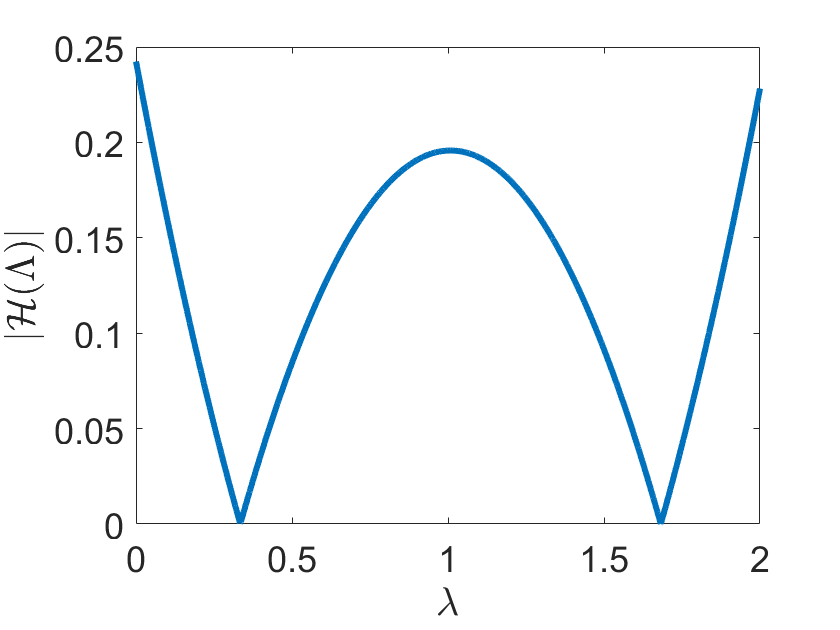}
        \caption{}
    \label{fig:citefilt}
    \end{subfigure}
    \begin{subfigure}[b]{0.19\linewidth}
    \includegraphics[width=\linewidth]{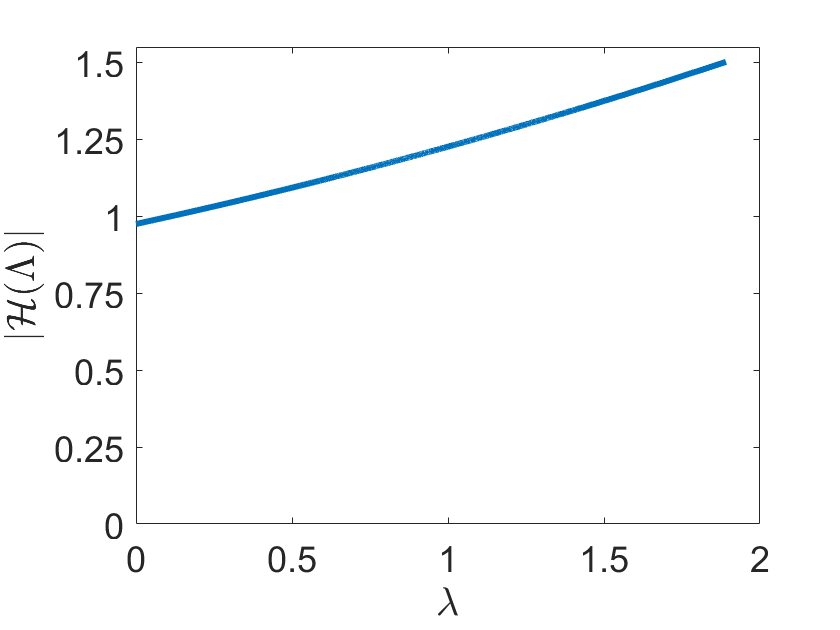}
        \caption{}
    \label{fig:sinafilt}
    \end{subfigure}
    \begin{subfigure}[b]{0.19\linewidth}
    \includegraphics[width=\linewidth]{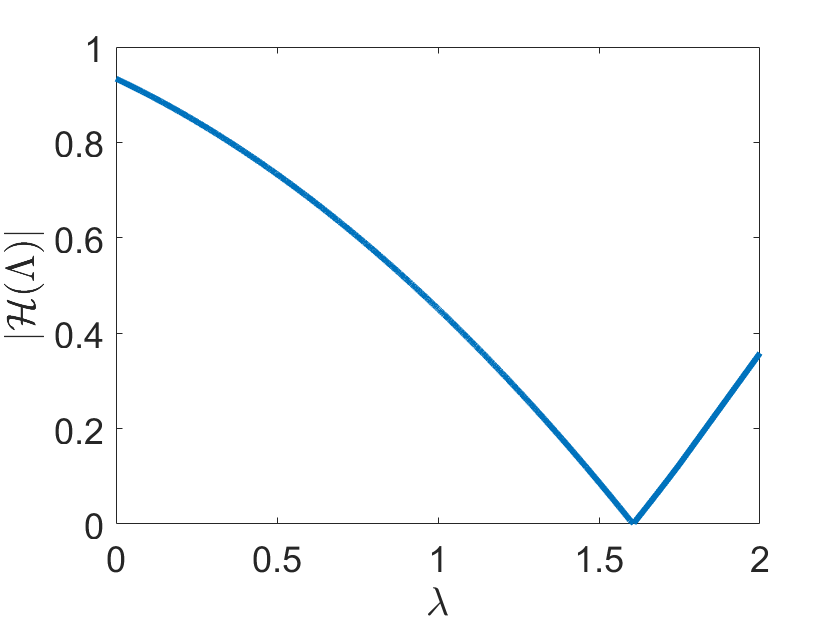}
        \caption{}
    \label{fig:wikifilt}
    \end{subfigure}
    \begin{subfigure}[b]{0.19\linewidth}
    \includegraphics[width=\linewidth]{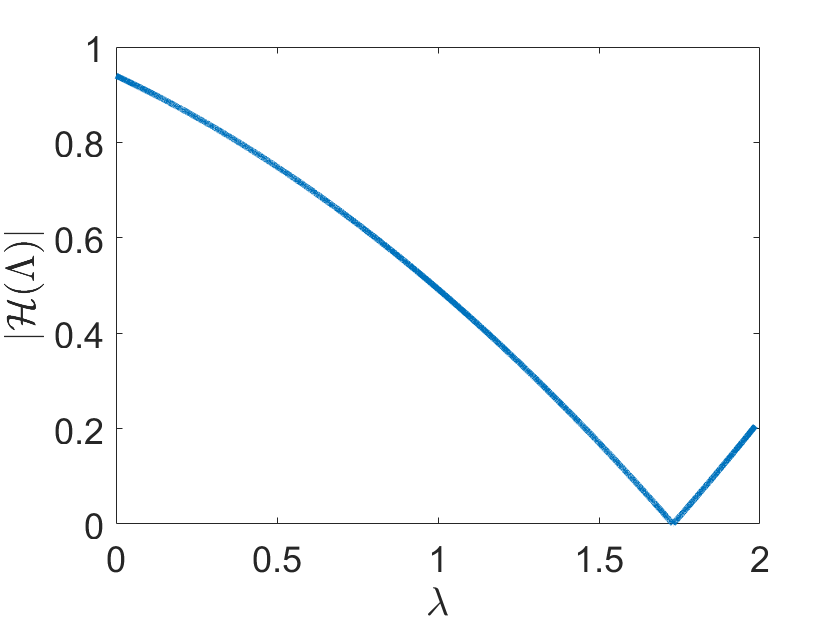}
    \caption{}
    \label{fig:pubfilt}
    \end{subfigure}
    \begin{subfigure}[b]{0.19\linewidth}
    \includegraphics[width=\linewidth]{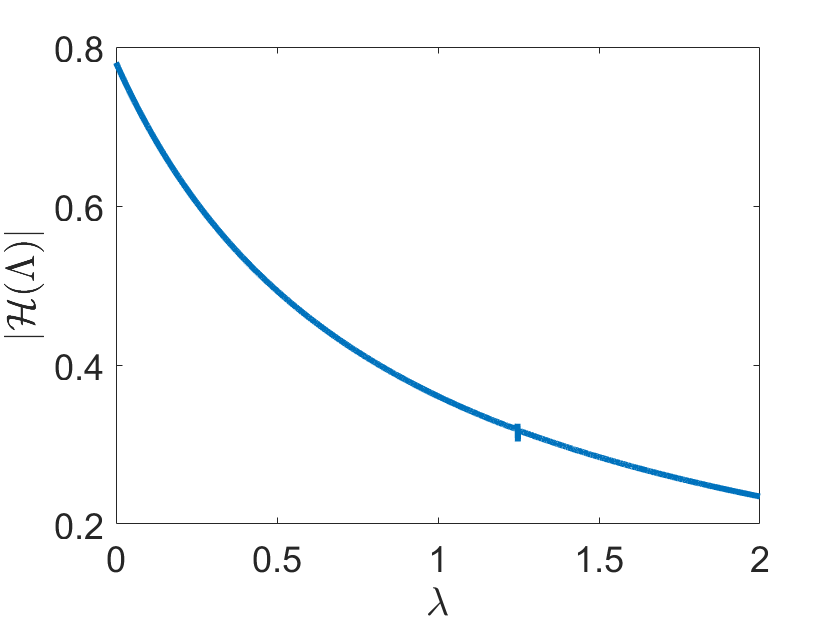}
        \caption{}
    \label{fig:corafiltARMA}
    \end{subfigure}
    \begin{subfigure}[b]{0.19\linewidth}
    \includegraphics[width=\linewidth]{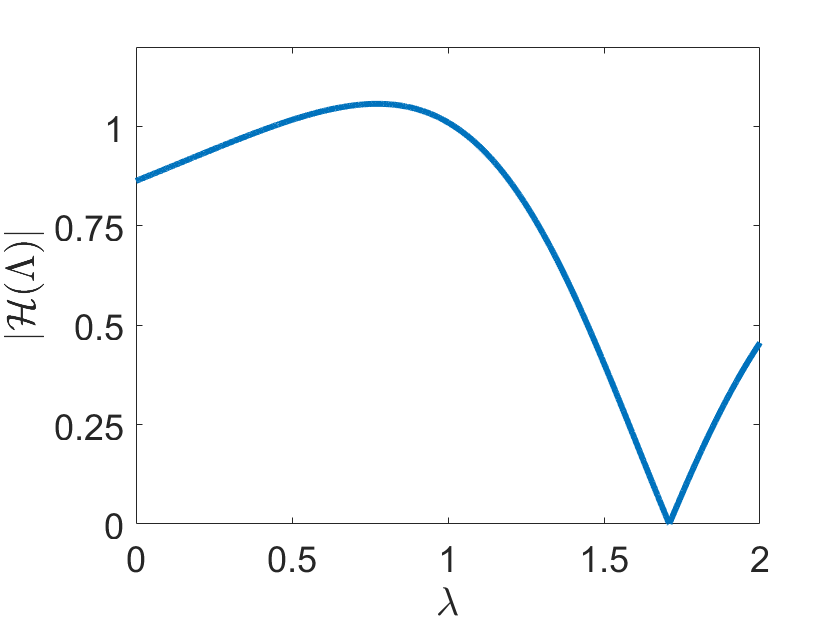}
        \caption{}
    \label{fig:citefiltARMA}
    \end{subfigure}
    \begin{subfigure}[b]{0.19\linewidth}
    \includegraphics[width=\linewidth]{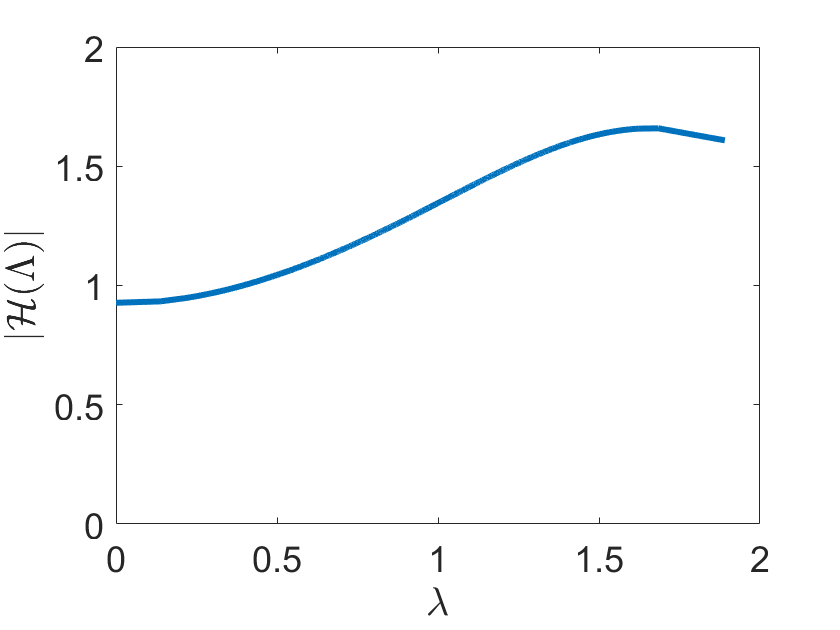}
        \caption{}
    \label{fig:sinafiltARMA}
    \end{subfigure}
    \begin{subfigure}[b]{0.19\linewidth}
    \includegraphics[width=\linewidth]{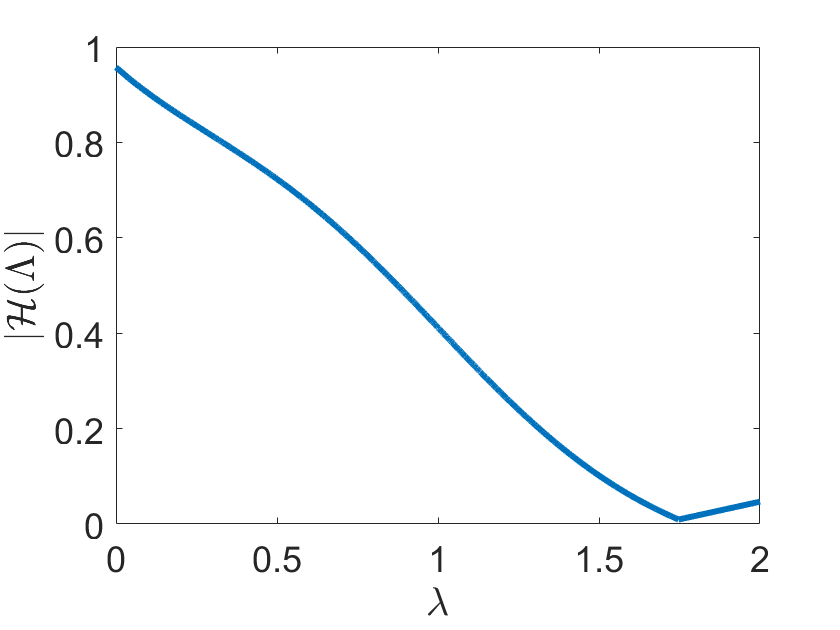}
        \caption{}
    \label{fig:wikifiltARMA}
    \end{subfigure}
    \begin{subfigure}[b]{0.19\linewidth}
    \includegraphics[width=\linewidth]{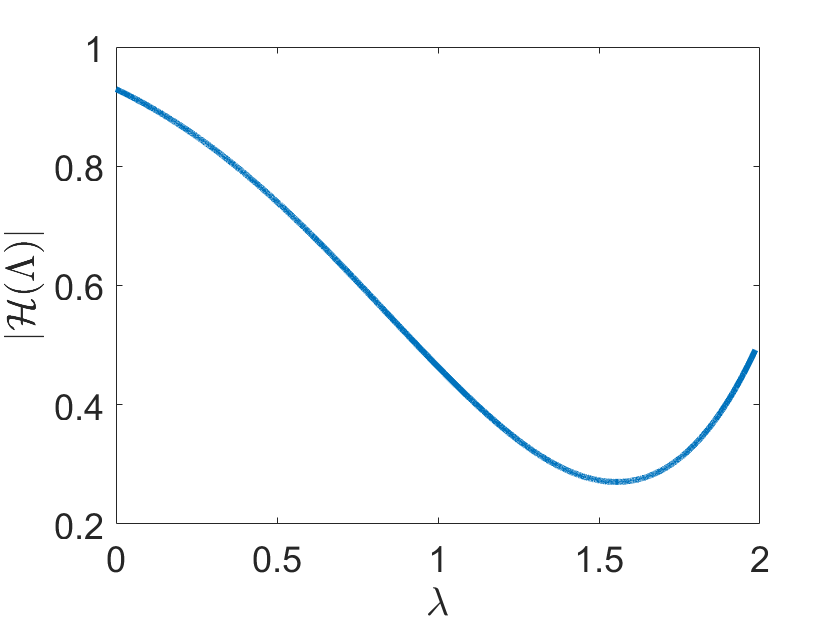}
    \caption{}
    \label{fig:pubfiltARMA}
    \end{subfigure}
    \caption{(a)-(e) Spectral Energy Distributions (SED) of the graph signal $\matr{F}$. (f)-(j) Optimal FIR graph filters for each dataset. (k)-(o) Optimal ARMA graph filters for each dataset. From left to right, Cora, Citeseer, Sinanet, Wiki, and PubMed, respectively. }
    \label{fig:filters}
\end{figure*} 

Table \ref{tab:results} shows the performance of all the methods. For all datasets, our method performs better than the state-of-the-art methods in terms of NMI. For PubMed, AGC  performs the best in terms of ARI, but GraFiCA performs the best in terms of NMI. For Citeseer and Wiki, EGAE performs better than GraFiCA with FIR in terms of ARI, but both GraFICA performances are higher in terms of NMI. For most datasets, the performance of the FIR and ARMA filters are very similar to each other. In the next section, we will discuss the interpretation of these results in terms of the learned filters. 

\subsection{Interpretation of the Learned Filters}

\begin{figure}[h]
    \vspace{-0.1in}
    \centering
    \includegraphics[width=0.42\linewidth]{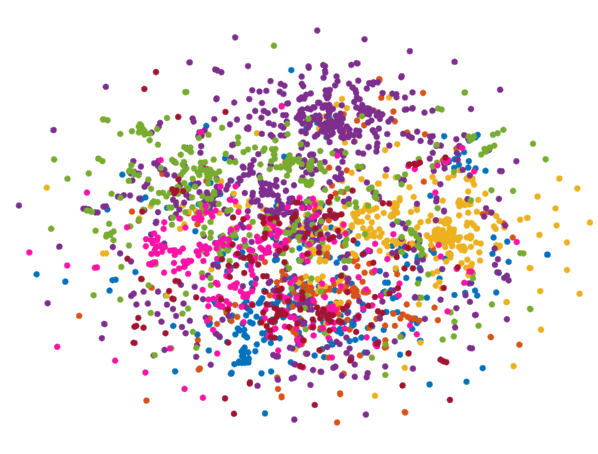}
    \includegraphics[width=0.4\linewidth]{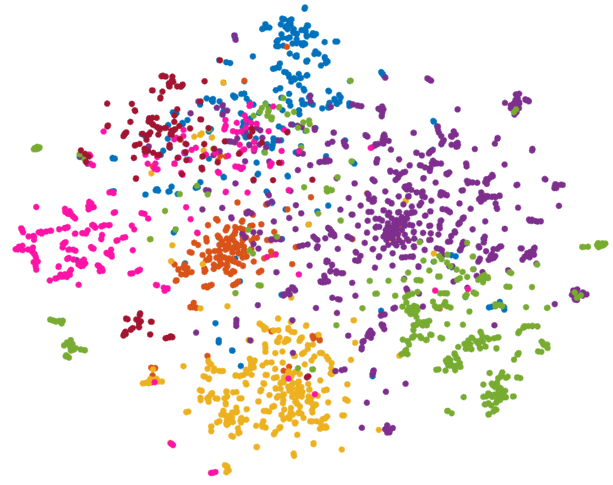}
    \\
    \includegraphics[width=0.38\linewidth]{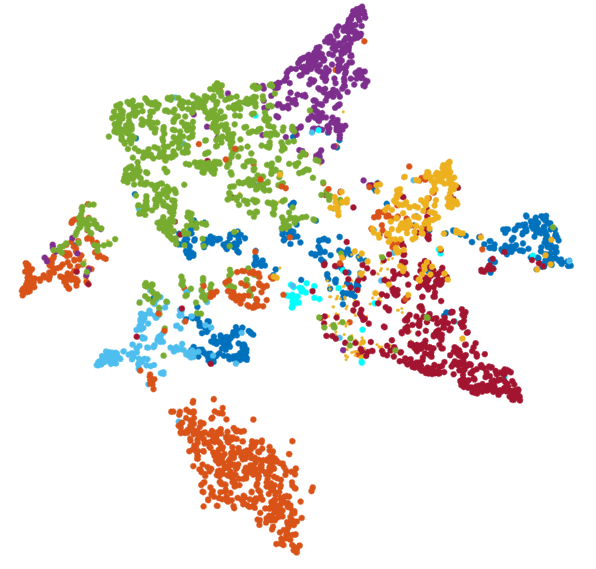}
    \includegraphics[width=0.33\linewidth]{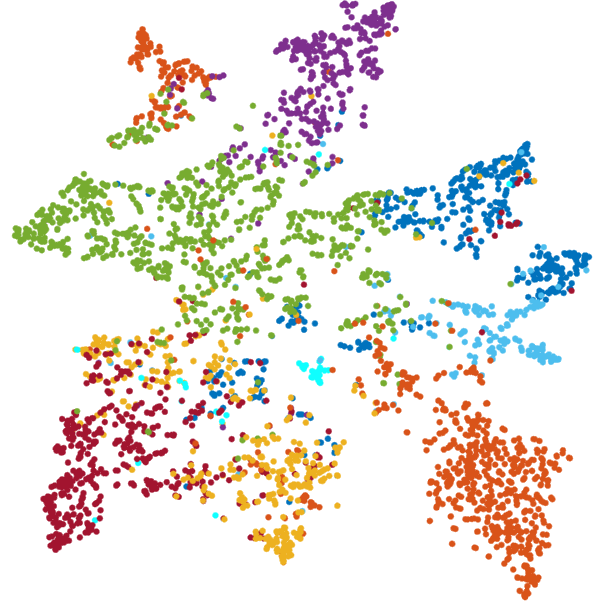}
    \caption{t-SNE visualization \cite{van2008visualizing} of the original (left) and filtered (right) attributes for Cora (top row) and Sinanet (bottom row). Each color represents a cluster.}
    \label{fig:att}
\end{figure}

Fig. \ref{fig:corafilt}-\ref{fig:pubfiltARMA} show the frequency responses of the optimal FIR and ARMA filters for each dataset. For most datasets, the optimal filters have both a low-pass and high-pass region, thus extracting both smooth and non-smooth features from the node attributes. This is in contrast to GAE-based methods that are limited to first-order low-pass filtering and AGC, which is a higher-order low-pass filter. While the clustering performance of FIR and ARMA filters are similar to each other, the filter shapes obtained by ARMA are smoother as they fit IIR filters to approximate the same frequency response. It is also interesting to note that the optimal filter for Sinanet is an all-pass filter, as the original node attributes carry most of the class information. In this case, filtering the attributes may not improve the accuracy of clustering as seen in the second row of Fig. \ref{fig:att} where the attributes before and after filtering are very similar. This also explains why methods like k-means and CGFKM which only rely on the attributes perform well in clustering this dataset. On the contrary, for Cora, the clusters become better separated after filtering, as seen in the first row of Fig. \ref{fig:att}. This is in line with the poor clustering performance of methods that only use node attributes.  
These results indicate that our method adapts to the characteristics of the data and yields interpretable filters. 

Except for the filters for Sinanet, which are all-pass, for most of the other datasets, the filters are mostly low-pass with content in the middle and high frequencies. The fact that our filters do not entirely suppress high-frequency components suggests a balance between preserving the cohesiveness of clusters and highlighting crucial differences across various regions of the graph.

In order to better understand and interpret the frequency responses of the learned filters, we examine the Spectral Energy Distribution (SED) of the graph signals with respect to the eigenvalues of the graph Laplacian.
For a graph signal $\matr{F}$ and its graph Fourier transform $\hat{\matr{F}}=\matr{U}^\top\matr{F}$, the spectral energy distribution at $\lambda_i$ is defined as the average of $\hat{F}_{ip}^2/\sum_{i=1}^N (\hat{F}_{ip})^2$ across the $p$ attributes.
Fig. \ref{fig:coraSED}-\ref{fig:pubSED} show the spectral energy distribution of the graph signals for each dataset. A concentration of energy in the lower end of the spectrum suggests that the graph signal is smooth and varies slowly across the graph. This can also indicate strong connectivity within certain graph regions or the presence of well-defined communities. 
Significant energy in the higher frequencies indicates that the graph signal has high variability or rapid changes across edges. This might also suggest that the graph contains regions of sparse connectivity and other regions with higher density. 

As we can see in Fig. \ref{fig:coraSED} and \ref{fig:citeSED}, Cora and Citeseer are similar in terms of their SED. They both have SED uniformly distributed across all frequencies with increased SED in the low frequencies.  Both datasets represent papers in different machine learning areas, where it is fair to assume that the papers could be easily related to more than just one of those areas and therefore, the clusters are not very well defined. Due to this distribution of SED, the optimal filters for Cora and Citeseer have significant power in the middle frequencies. On the other hand, in Sinanet, the main areas of the 10 different forums range from parenting, history and arts, politics,  to science and technology, and the clusters are well separated, hence there is a significant peak at the low frequencies as seen in Fig. \ref{fig:sinaSED}. As for the peak in the high frequencies for SED of Sinanet, this might be due to the different densities in the clusters. Wiki and Pubmed show similar behavior with respect to their SEDs, with most of the SED concentrated in the lowest frequencies  as seen in Fig. \ref{fig:filters}. Similar to their SED profiles, the shape of the optimal filters for these two datasets are similar and primarily lowpass.





\subsection{Parameter Sensitivity}
\begin{figure}[h]
    \centering
    \includegraphics[width=0.48\linewidth]{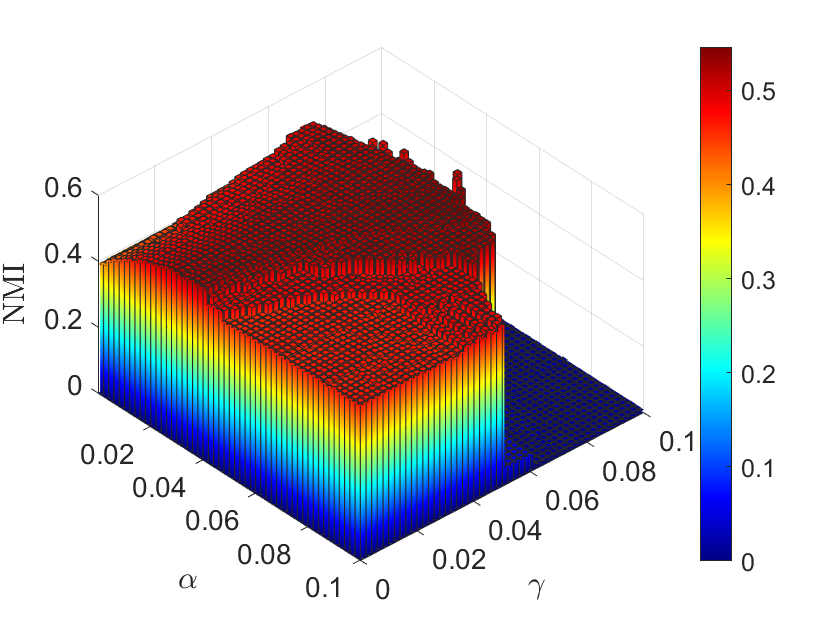}
    \includegraphics[width=0.48\linewidth]{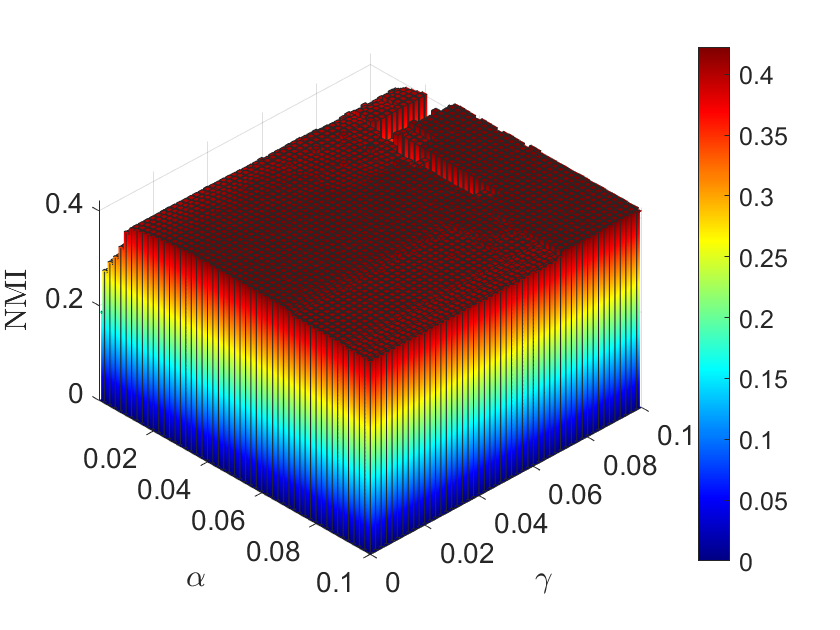} \includegraphics[width=0.48\linewidth]{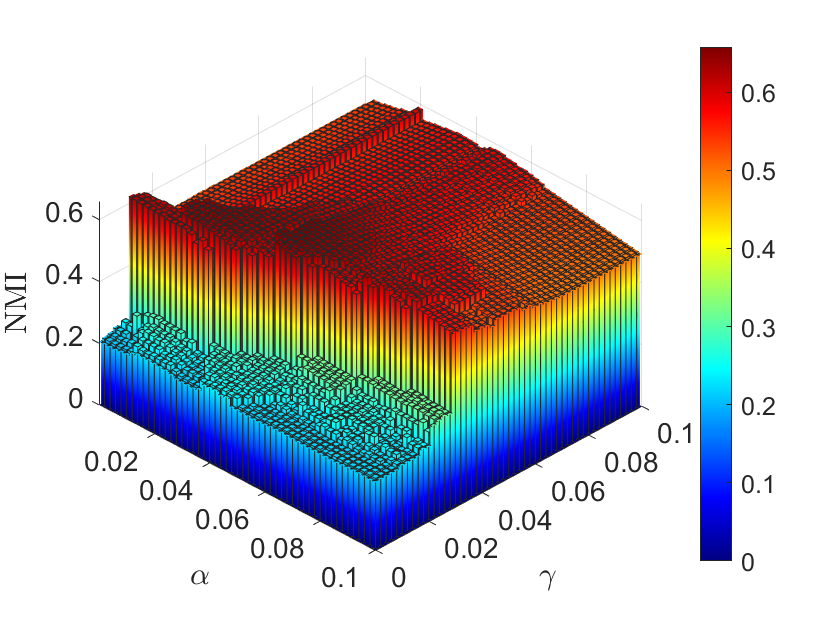}
    \includegraphics[width=0.48\linewidth]{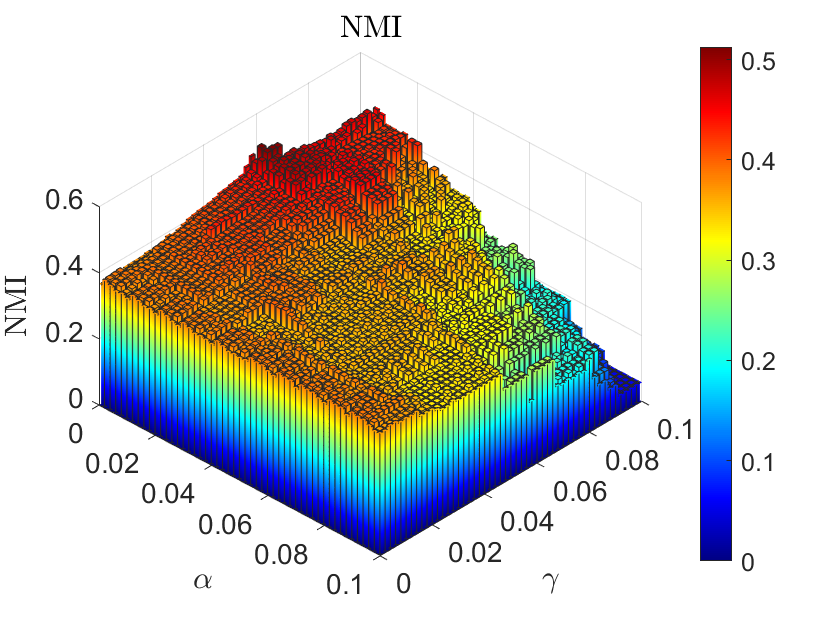}
    \includegraphics[width=0.48\linewidth]{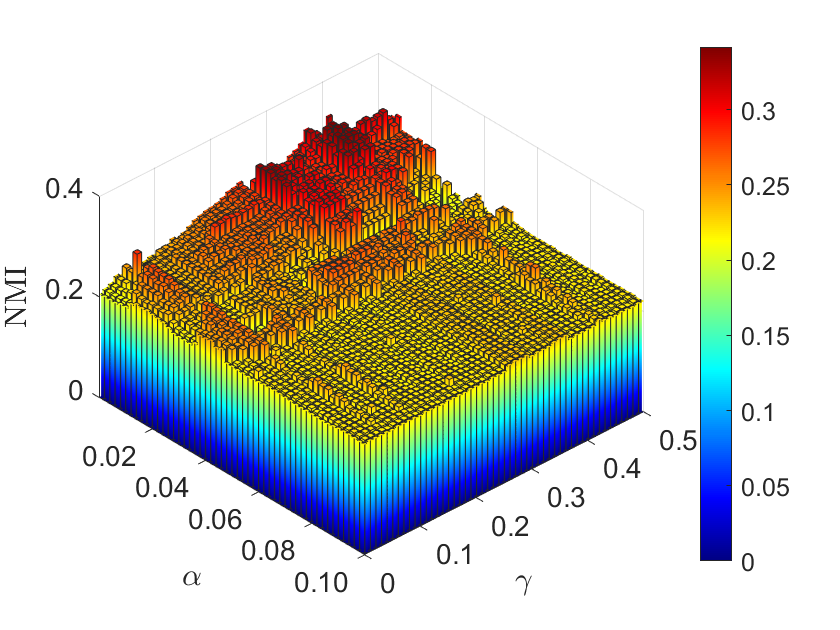}
    \caption{Parameter sensitivity for Cora, Citeseer, Sinanet, Wiki, and PubMed with $T=3$.}
    \label{fig:par}
\end{figure}

We investigated the effect of the two hyperparameters, $\alpha$ and $\gamma$, on clustering accuracy for FIR filters for $T=3$ as seen in Fig. \ref{fig:par}. For sparse networks such as Cora and Citeseer, the connectivity information does not help to improve the performance, thus making the performance invariant to the choice of $\alpha$. 
On the other hand, for dense networks such as Wiki, the performance is more sensitive to the choice of $\alpha$. Finally, for Sinanet where most of the community information is reflected by the attributes, the performance is sensitive to the choice of $\gamma$  as it quantifies the tradeoff between within and between class association of the filtered attributes.    

\begin{figure}
    \centering
    \includegraphics[width=0.95\linewidth]{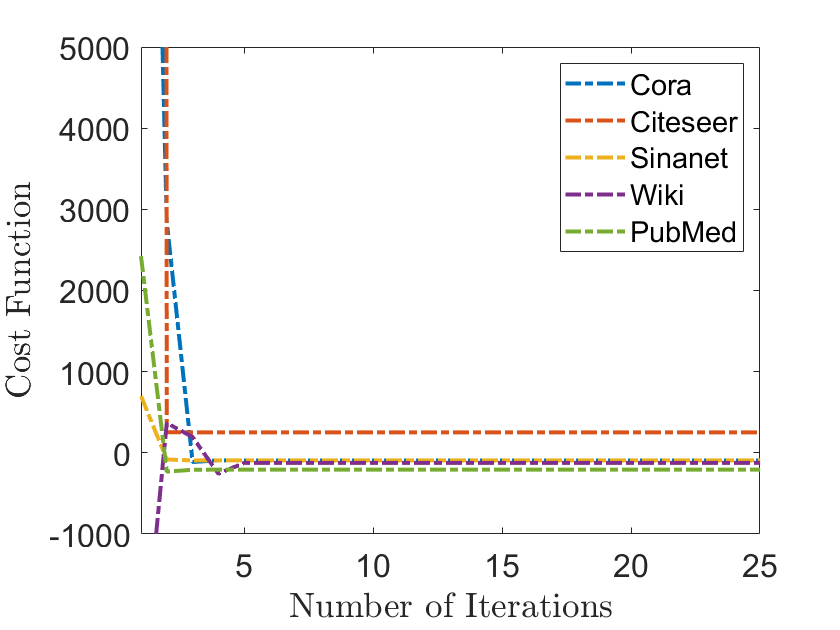}
    \caption{Cost function value vs the number of iterations}
    \label{fig:conv}
\end{figure}
\vspace{-0.1in}

\begin{figure*}[h]
    \centering
    \includegraphics[width=0.19\linewidth]{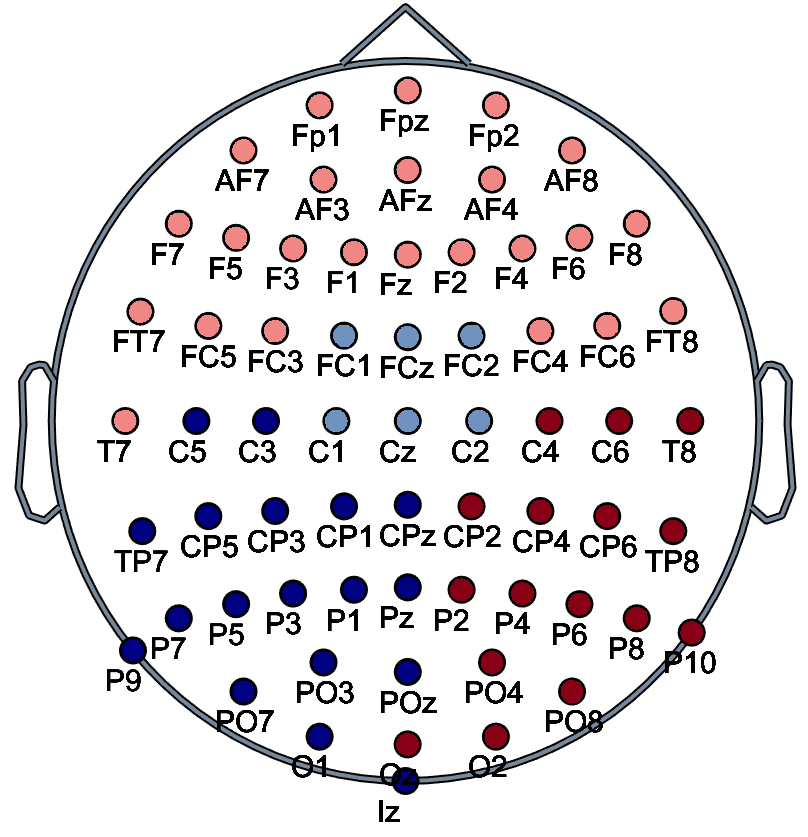}
    \includegraphics[width=0.19\linewidth]{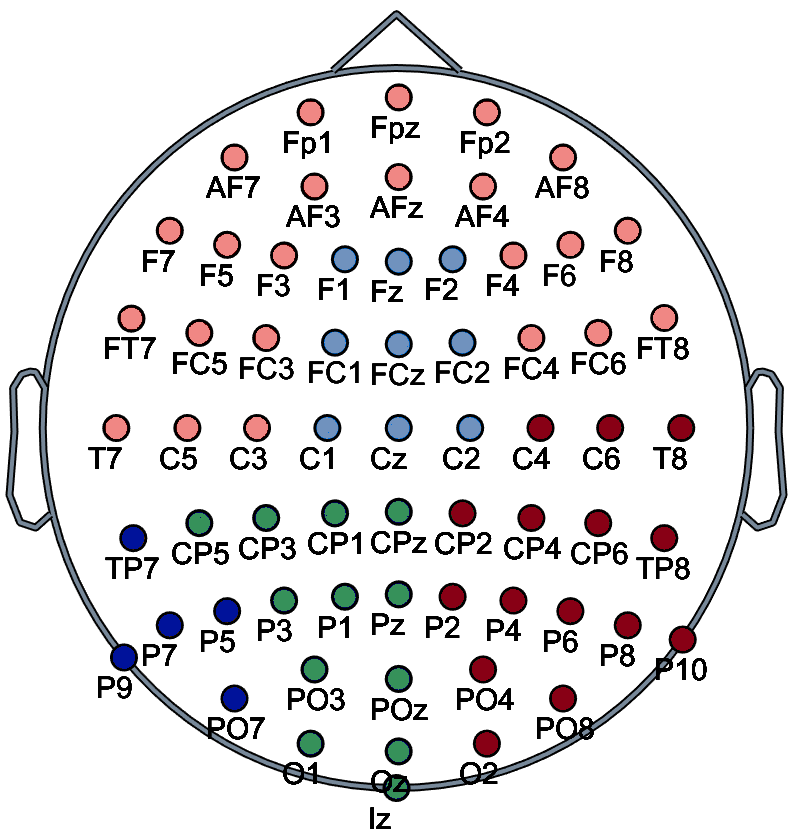}
    \includegraphics[width=0.19\linewidth]{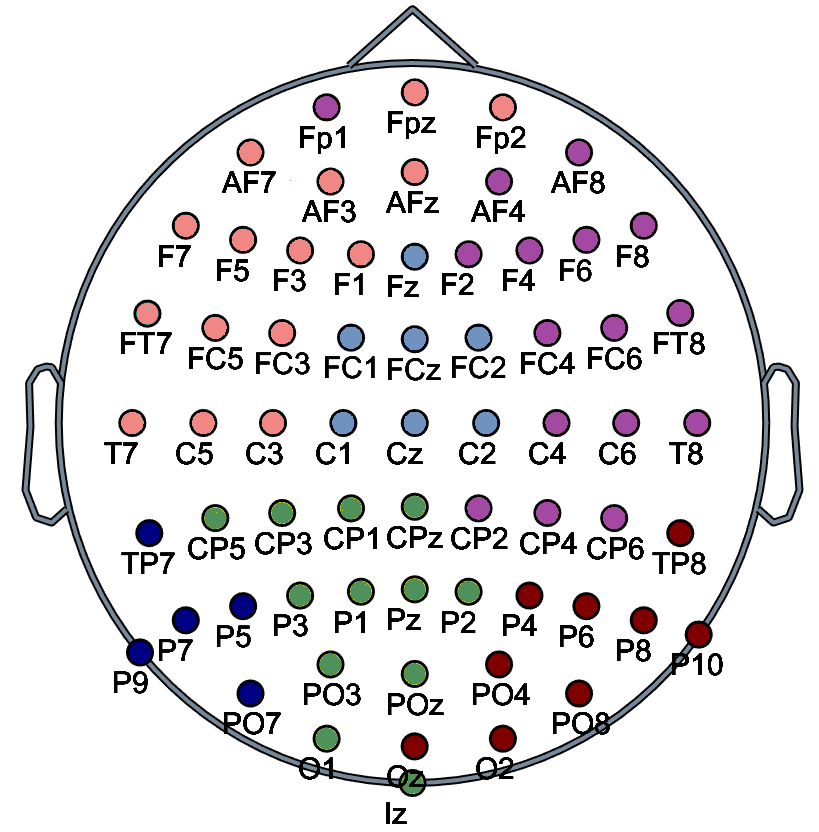}
    \includegraphics[width=0.19\linewidth]{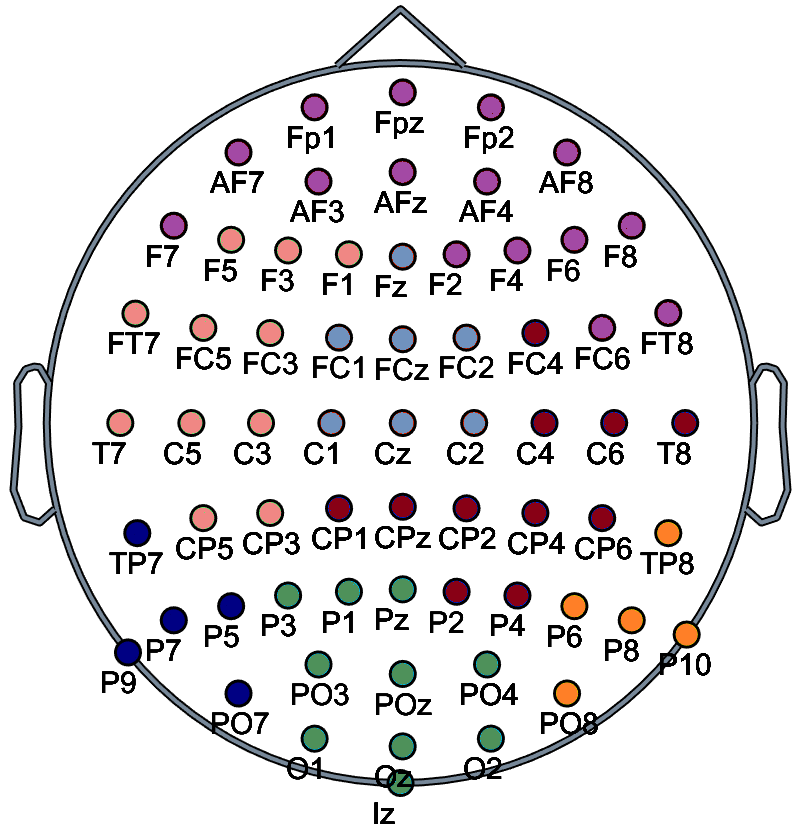}
    \includegraphics[width=0.19\linewidth]{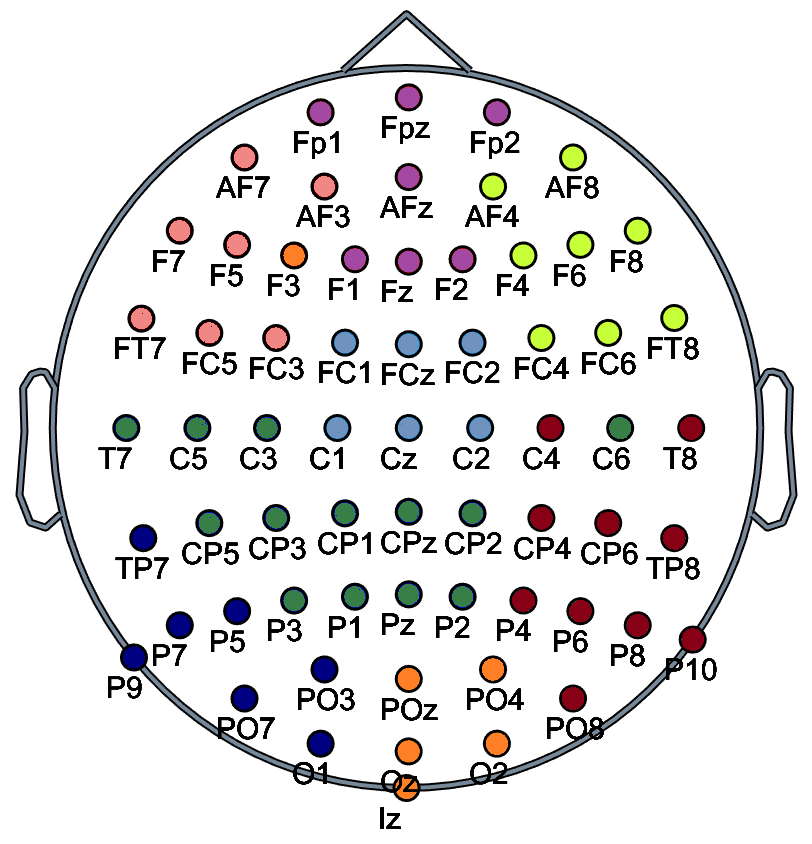}
    \caption{Community structure for $K=4, 5, 6, 7, 8$}
    \label{fig:eegk}
\end{figure*}  
\subsection{Convergence Analysis}
The optimization problem in \eqref{eq:optprob} is solved in an alternating way, i.e., we fix one variable and optimize the other. When $\matr{h}$ is fixed, the solution of \eqref{eq:probstep1}, $\matr{\bar{Z}}$, obtained by selecting the  $K$ eigenvectors corresponding to $K$ smallest eigenvalues of $\matr{\tilde{W}}_{n}-2\alpha\matr{A}_{n}$ is the global optimum solution to the problem in \eqref{eq:probstep1}. Similarly, when $\matr{\bar{Z}}$ is fixed, i.e., the community structure is known, the solution of \eqref{eq:probsteph} $\matr{h}$ is the eigenvector corresponding to the smallest eigenvalue of $\matr{B}_T-\gamma\matr{C}_T$ and is a global solution to the problem in \eqref{eq:probsteph}. Although $\matr{\bar{Z}}$ is the global optimum solution to \eqref{eq:probstep1}, the partition $\boldsymbol{\mathcal{C}}$ is found by applying $k$-means to $\matr{\bar{Z}}$. While k-means  is known to converge quickly, it does not guarantee convergence to a global optimum. Thus, while each step of the algorithm converges to a global optimum, the final partition may not be globally optimum.

Fig. \ref{fig:conv} illustrates the empirical convergence behavior, where the value of the cost function as a function of the number of iterations is given for the different datasets. Despite the local nature of k-means convergence, our overall algorithm consistently converges within a few number of iterations.

\subsection{Application to Brain Functional Connectivity Network}

We applied the proposed attributed graph clustering framework to functional connectivity networks of the brain. Electroencephalogram (EEG) data collected from a cognitive control-related error processing study \cite{hall2007externalizing}, i.e., Flanker task, was used to construct both the graphs and the graph signals.  The EEG was recorded following the international 10/20 system for placement of 64 Ag–AgCl electrodes. The sampling frequency was 512 Hz. After the removal of the trials with artifacts, the Current Source Density (CSD) Toolbox 
\cite{tenke2012generator} was employed to minimize the volume conduction.  
In this study, trials corresponding to Error-Related Negativity (ERN) after an error response were used. Each trial was one second long. The total number of trials was 480 in which the total number of error trials in different participants varied from 20 to 61.  

As previous studies indicate neural oscillations in the theta-band (4–7 Hz) may be one mechanism that underlies functional communication between networks involving medial prefrontal cortex (mPFC) and lateral prefrontal cortex (lPFC) regions during the ERN (25–75 ms time window) \cite{hall2007externalizing,trujillo2007theta,cavanagh2014frontal,bolanos2013weighted}, all analysis was 
performed for this time and frequency range. The average phase synchrony corresponding to theta band and 25-75 ms time window were computed to construct 64 $\times$ 64 connectivity matrices for each subject. The graph signal for subject $l$, $\matr{F}^l\in \mathbb{R}^{64 \times 512}$, is defined as the average time series across trials for each electrode. In this paper, we consider data from 20 participants. The FCNs across subjects can be modeled as a multiplex network with  64 nodes and 20 layers, corresponding to the number of brain regions and subjects, respectively. 
\subsubsection{GraFiCA Implementation}

The proposed method, GraFiCA, is extended to multiplex networks with $L$ layers with adjacency matrix $\matr{A}^{l}\in \mathbb{R}^{N\times N}$ and the corresponding graph signal, $\matr{F}^{l}\in \mathbb{R}^{N\times P}$ for the FIR filter. The consensus community structure across layers can be learned by extending the proposed cost function in \eqref{eq:optprob} as
\begin{equation}
\begin{split}
 \sum_{l=1}^{L}\sum_{k=1}^{K}\frac{1}{\text{vol}(\mathcal{C}_k)^l}\sum_{i,j\in \mathcal{C}_k}||\tilde{F}_{i\cdot}^l-\tilde{F}_{j\cdot}^l||^2\\-\gamma\sum_{l=1}^{L}\sum_{k=1}^{K}\frac{1}{\text{vol}(\mathcal{C}_k)^l}\sum_{\substack{i\in \mathcal{C}_k\\
     j\notin \mathcal{C}_k}}||\tilde{F}_{i\cdot}^l-\tilde{F}_{j\cdot}^l||^2.
\end{split}
\label{eq:mxoptprob}
\end{equation}
\noindent The partition $\boldsymbol{\mathcal{C}}$ and the FIR filter coefficients $\matr{h}$ can be found by extending the corresponding optimization problems derived above as
\begin{equation}
\begin{split}
&\boldsymbol{\matr{\bar{Z}}}
:=\underset{\matr{\bar{Z}},\matr{\bar{Z}}^\top\matr{\bar{Z}}=\matr{I}}{\argmin} \hspace{0.1cm}\text{tr}(\matr{\bar{Z}}^\top\sum_{l=1}^L(\matr{\tilde{W}}_n^l-2\alpha\matr{A}^l_{n})\matr{\bar{Z}}),\\
&\matr{h}:=\underset{\matr{h}}{\argmin}\hspace{0.2cm} (\matr{h}^\top\sum_{l=1}^L(\matr{B}^l-\gamma \matr{C}^l)\matr{h}).
\end{split}
\label{eq:optstepsL}
\end{equation}





Fig. \ref{fig:eegk} shows the multiplex community structure for different numbers of clusters, $K$, from 4 to 8 across 20 subjects, and $T=3$. As shown in Fig. \ref{fig:eegk}, for each value of $K$, we consistently obtain a  community comprised of the frontal-central
nodes corresponding to the medial prefrontal cortex (mPFC), e.g. FCz, FCz, FC2.  Frontal-central connectivity in
theta oscillations is known to play a critical role in the flexible management of cognitive control \cite{mcloughlin2022midfrontal}. In addition, the community structure reveals distinct communities corresponding to the visual and lateral prefrontal cortex (lPFC) areas. 

\begin{figure}[h]
    \centering
    \includegraphics[width=0.49\linewidth]{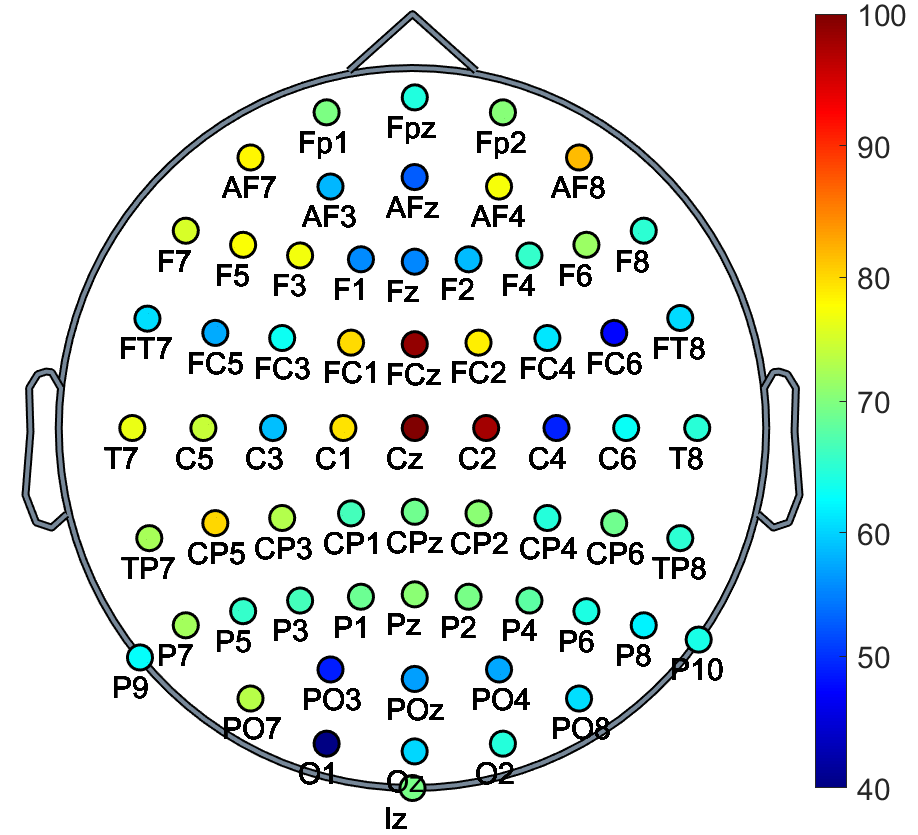}
    \includegraphics[width=0.49\linewidth]{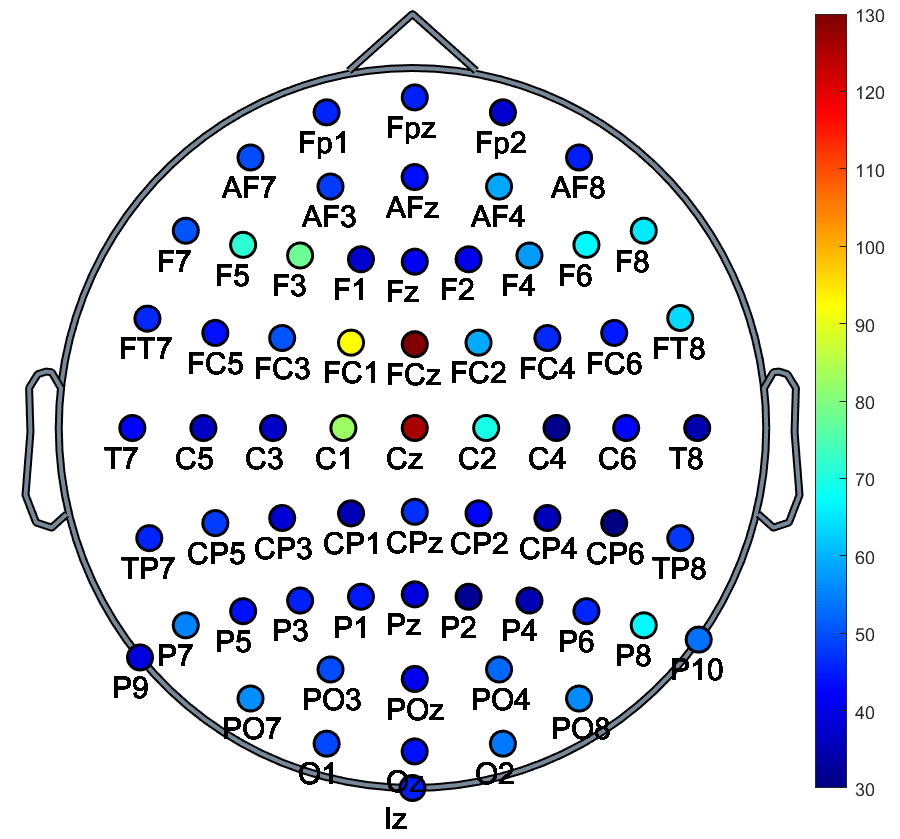}
    \caption{Global Scaled Inclusivity}
    \label{fig:GSI}
\end{figure}

Next, we evaluated the consistency of the community structure obtained from the multiplex network with the community structure of individual subjects. We applied GraFiCA to each layer individually, and the optimal FIR filter with $T=3$ and the corresponding community structure for each subject was learned with different values of $K$ ranging from 4 to 8. Scaled inclusivity (SI) \cite{steen2011assessing} was employed as a metric to evaluate the consistency of the community structure across subjects. SI is calculated by measuring the overlap of communities across multiple networks while penalizing for the disjunction of communities \cite{steen2011assessing,moussa2012consistency}. The Global Scaled Inclusity (GSI) \cite{steen2011assessing,moussa2012consistency} across these 20 community structures is calculated. Fig. \ref{fig:GSI} shows the GSI for $K=4$ and $K=6$. In both cases, the central nodes, FC1, FCz, FC2, C1, Cz, and C2, are among the 10 nodes with the highest GSI values. As we can see from Fig. \ref{fig:eegk}, these 6 nodes are consistently detected in the same community which indicates that the multiplex extension of the algorithm obtained communities that are consistent with the individual subjects' community structure.

\section{Conclusions}
\label{sec:conclusions}
As the amount of large-scale network data with node attributes increases, it is important to develop efficient and interpretable graph clustering methods that identify the node labels. In this paper, we introduced an optimal parametric graph filter learning method for attributed graph clustering. The proposed method makes some key contributions to the field. First, the parameters of polynomial graph filters are learned with respect to a loss function that quantifies both the association within and the separation between clusters. Thus, the filter parameters are optimized for the clustering task.
Second, the proposed method does not constrain the filters to be lowpass. Results indicate that the learned filters take into account the useful information in middle and high-frequency bands and the structure of the filter is determined directly by the data. Third, the method is formulated for both FIR and IIR filters, providing similar performance across datasets. While FIR filters are computationally less expensive, IIR filters provide a smoother frequency response. Finally, the learned filters are evaluated with respect to spectral energy distribution of the attributed graphs providing interpretability to the proposed design procedure. 
Future work will consider learning a combination of filters and extension to nonlinear filters such as graph wavelets. 

\bibliographystyle{IEEEbib}
\bibliography{refs}

\end{document}